\documentclass[preprint,12pt]{elsarticle}
\usepackage[utf8]{inputenc} % allow utf-8 input
\usepackage[T1]{fontenc}    % use 8-bit T1 fonts
\usepackage{hyperref}       % hyperlin/ks
\usepackage{url}            % simple URL typesetting
\usepackage{booktabs}       % professional-quality tables
\usepackage{microtype}      % microtypography
\usepackage{lipsum}	    	% Can be removed after putting your text content /
\usepackage{multirow}
\usepackage{graphicx}
\usepackage{doi}
\usepackage[section]{placeins}
\usepackage{mathtools}
\usepackage{amsmath}
\usepackage{amssymb}
\usepackage[ruled,vlined]{algorithm2e}
\usepackage{lineno}
\usepackage[dvipsnames]{xcolor}
\usepackage{enumitem}
\usepackage{tabularx} % Required for adjusting table width
\usepackage{array}
\usepackage{subcaption}

\usepackage{siunitx}

\usepackage{{makecell}}
\setcellgapes{4pt} % Adjust the cell padding

\newcolumntype{Y}{>{\centering\arraybackslash}X}
\newcolumntype{M}[1]{>{\centering\arraybackslash}m{#1}}

\setlist[itemize]{noitemsep, topsep=0pt}
\usepackage{amssymb}
\usepackage[a4paper, total={6in, 8in}, margin=1in]{geometry}

\journal{Elsevier}
\usepackage{color}
\usepackage{soul}

\begin{document}

\begin{frontmatter}

\title{Physics-informed Multiple-Input Operators for efficient dynamic response prediction of structures}

\author[inst1]{Bilal Ahmed\corref{cor1}}\ead{ba2702@nyu.edu}
\author[inst1,inst2]{Yuqing Qiu}
\author[inst1,inst3]{Diab W. Abueidda\corref{cor1}}\ead{da3205@nyu.edu}
\author[inst1,inst4]{Waleed El-Sekelly}
\author[inst1]{Tarek Abdoun}
\author[inst1]{Mostafa E. Mobasher\corref{cor1}}\ead{mostafa.mobasher@nyu.edu}

\affiliation[inst1]{
    organization={Civil and Urban Engineering Department},
    addressline={New York University Abu Dhabi}, 
    country={United Arab Emirates}
}

\affiliation[inst2]{
    organization={The State Key Laboratory of Mechanics and Control of Mechanical Structures},
    addressline={Nanjing University of Aeronautics and Astronautics}, 
    country={China}
}

\affiliation[inst3]{
    organization={National Center for Supercomputing Applications},
    addressline={University of Illinois at Urbana-Champaign}, 
    country={United States of America}
}

\affiliation[inst4]{
    organization={Department of Structural Engineering},
    addressline={Mansoura University}, 
    country={Mansoura, Egypt}
}

\cortext[cor1]{Corresponding author}

\begin{abstract}

Finite element (FE) modeling is crucial for structural analysis but remains computationally expensive, particularly for dynamic loading scenarios. Recently, operator learning models have successfully replicated structural responses at FEM level under static loading \citep{ahmed2025physics}; however, modeling dynamic structural behavior remains a significantly more complex challenge. In this work, we address this problem by developing a multiple-input operator network (MIONet) architecture that incorporates a second trunk network to explicitly encode temporal dynamics, thereby enabling the accurate prediction of structural responses under moving loads. Traditional deep operator networks (DeepONet) architectures, particularly those using recurrent neural networks (RNNs), rely on fixed-time discretization, which limits their ability to capture continuous dynamic responses in real-world discrete structures. In contrast, the proposed MIONet architecture enables seamless prediction across both space and time, eliminating the need for step-wise or sequential modeling. This allows the network to map scalar inputs, including moving load parameters, velocity, spatial discretization, and time steps, to a continuous structural response. To enhance efficiency and ensure physical consistency, we introduce novel physics-informed learning, which leverages a precomputed mass, damping, and stiffness matrix to enforce dynamic equilibrium without explicitly solving the governing partial differential equations (PDEs). Additionally, we employ the Schur complement to reduce computational cost by training the model in a reduced domain, significantly reducing training costs while maintaining accuracy across the entire structural domain. The proposed method is validated on both a simple beam structure and the real-world KW-51 bridge, demonstrating its ability to produce FEM-level accurate predictions within fractions of a second, making it suitable for applications requiring real-time predictions, such as digital twins. Comparative studies against GRU-DeepONet demonstrate that our approach achieves similar accuracy and ensures temporal continuity. Additionally, it delivers over 100-fold faster computation compared to conventional FEM simulations, offering a highly efficient alternative for dynamic structural analysis.
\end{abstract}

\begin{highlights}
\item Proposed MIONet for real-time structural response prediction under dynamic loading.
\item Outperforms GRU-based DeepONet in capturing continuous temporal dynamics.
\item Enforces physics via dynamic equilibrium matrix constraints.
\item Achieves 95\%+ accuracy and 100× speedup in dynamic response prediction.

\end{highlights}


\end{frontmatter}


\section{Introduction} \label{intro}

\subsection{Overview}\label{Intro_Overview}

Over the past few decades, FEM \cite{brenner2002mathematical,hughes2012finite} has been widely recognized as a common and robust tool for simulating complex engineering and mechanical systems by numerically solving PDEs, particularly in cases where analytical solutions are intractable or excessively complicated. Although FEM provides highly accurate approximations, its computational cost increases significantly as the complexity of the problem increases, particularly in real-world dynamic loading scenarios. The repeated time-stepping and re-solving required by FEM for dynamic loading makes it computationally expensive and inefficient for time-dependent problems. Moreover, even minor changes in loading or structural parameters require a complete reanalysis. This limitation becomes critical for real-time structural monitoring and digital twin applications \citep{chiachio2022structural,ye2019digital,ritto2021digital}, which increasingly require the ability to model and predict dynamic structural responses rapidly and accurately under varying loading conditions. Although our previous operator learning approach \citep{ahmed2025physics} demonstrated success for static loading scenarios, its applicability to truly dynamic loading and response predictions remains limited. To address this gap, we propose an MIONet architecture designed to predict full-field structural responses, namely shifts and rotations, in both space and time under varying moving loads and velocities. By incorporating a dual-trunk design, the framework ensures continuity in both domains while maintaining consistency with dynamic equilibrium principles. Once trained, the model enables real-time inference at a fraction of the computational cost of FEM, offering a robust and scalable surrogate for dynamic structural simulations in practical monitoring and digital twin scenarios.

\subsection{Literature Review and Research Gaps}\label{Intro_Lit}

Recent advances in scientific machine learning (ML) have demonstrated greater effectiveness in approximating the complex, dynamic behavior of systems compared to conventional numerical approaches. ML models can significantly reduce computational costs by addressing real-life problems across diverse fields, such as solid mechanics and structural health monitoring \cite{brodnik2023perspective, li2021attention, LIN2022114553, liu2021multiscale}. 

Structural dynamic analysis remains a fundamental challenge in civil engineering. Traditionally, it relies on mathematical formulations using differential equations, which are solved through numerical methods such as the FEM \cite{zienkiewicz2005finite}, combined with time integration schemes like Newmark’s Beta method \cite{newmark1959method}. While effective, transitioning from static to dynamic analysis introduces additional numerical assumptions through these schemes, often resulting in challenges related to stability, convergence, and computational efficiency.With the rise of data-driven methods, recent research has explored learning time-sequential structural responses using deep learning architectures such as Recurrent Neural Networks (RNNs) \cite{elman1990finding}, Long Short-Term Memory (LSTM) networks \cite{graves2012long}, WaveNet \cite{oord2016wavenet}, and Residual Networks (ResNet) \cite{qin2019data}. Among these, LSTMs have demonstrated strong performance in predicting structural responses under seismic excitations \cite{kim2020development, ahmed2023unveiling, ahmed2022seismic, ahmed2023generalized}. Similarly, ResNet-based models have shown effectiveness in approximating both linear and nonlinear dynamical behaviors \cite{fu2020learning}. More recently, attention-based LSTM models have been employed to improve temporal learning, such as the attention-enhanced LSTM used by Liao et al. \cite{liao2023attention} to predict mid-span deflections of a cable-stayed bridge under seismic loading. In parallel, Oh et al. \cite{oh2023time} applied Convolutional Neural Networks (CNNs) to estimate long-term strains in structures arising from dead loads, temperature changes, creep, and shrinkage. Jiang et al. \cite{jiang2022structural} further extended the use of data-driven techniques by integrating Multi-Layer Perceptrons (MLPs) with particle swarm optimization to predict vertical displacements at sensor-specific locations. A wide body of literature also exists focusing on predicting mid-span deflections, strains, displacements, damage states, and inter-story drifts using neural networks and deep learning techniques \cite{radovanovic2015prediction, qiu2024damage, tadesse2012neural, abdu2023assessment, huang2021deep, kumar2021rapid}. While these studies present valuable contributions, most are limited to discrete spatial predictions—typically at sensor locations—and are purely data-driven. This limitation hinders their ability to ensure spatiotemporal consistency and generalize to unobserved regions of the structure. To address these gaps, the current study introduces a physics-informed learning framework for structural dynamics. This approach enables the prediction of full-field structural responses over time under dynamic loading, embedding governing physical laws and ensuring spatiotemporal consistency critical to dynamic structural behavior.

A significant challenge in artificial neural networks is their limited generalization capability, requiring retraining or adjustments when faced with changes in input parameters, discretization, or out-of-domain data. This limitation arises because traditional neural networks map inputs to outputs without explicitly learning the underlying physical phenomena. To address this, Chen et al. \cite{chen1995universal} introduced operator learning, which enables models to learn mappings between function spaces, allowing them to capture fundamental relationships in functional domains. Building on this, Lu et al. \cite{lu2021learning} proposed DeepONet, an operator learning framework capable of training with limited datasets while minimizing generalization errors. DeepONet employs a branch network to encode the input function and a trunk network to represent the output domain, enabling it to learn mappings between function spaces and solve families of PDE-related problems, including integrals and derivatives. Several extensions of DeepONet have since been developed, including Bayesian DeepONet \cite{lin2021accelerated,moya2023bayesian}, DeepONet with proper orthogonal decomposition \cite{lu2022comprehensive}, multiscale DeepONet \cite{liu2021multiscale}, neural operators with coupled attention \cite{kissas2022learning}, physics-informed DeepONet \cite{wang2021learning, moya2023dae}, and multiple-input deep neural operators (MIONet) \cite{jin2022mionet}. MIONet extends the DeepONet framework by introducing multiple parallel branch networks, enabling the model to handle multiple input functions simultaneously. Operator learning techniques have demonstrated strong potential for learning solutions, particularly for continuous domains, with applications in continuum mechanics, including advection, Burgers' equation, diffusion, and wave propagation \cite{wang2021learning, HE2024107258, koric2023data, yin2022interfacing, goswami2023physics, he2023novel}. However, to the best of the authors' knowledge, no prior studies have applied operator learning to predict dynamic responses in discrete structural systems under dynamic loading. While the authors have previously applied DeepONet to discrete structures for static loading \cite{ahmed2025physics}, capturing temporal dependencies in this domain remains a significant challenge. This is particularly crucial for real-time predictions of dynamic structural responses, where accurate temporal modeling is essential.

Several studies have attempted to address the challenge of capturing temporal dependencies in operator-based models. To incorporate temporal dynamics, researchers have investigated hybrid approaches that integrate DeepONet with RNNs. One such approach, Sequential DeepONet (S-DeepONet) \cite{he2024sequential, he2024predictions}, incorporates an RNN to process time-dependent inputs, improving accuracy in transient problems. However, this method relies on fixed-time windows and initial conditions, making it unsuitable for achieving truly continuous predictions in time. Another approach, proposed by Michalowska et al. \cite{michalowska2024neural}, employs a two-step method where a standard DeepONet first processes the input, followed by an RNN that post-processes the outputs using a moving window technique. Similarly, Bayesian MIONet with LSTM (B-LSTM MIONet) \cite{kong2023b} has been explored for learning variable time-dependent data, demonstrating effectiveness in capturing temporal dependencies. While these methods have proven successful for generalized PDEs—such as the Lorenz 63 system \cite{tabor1981analytic}, pendulum swing-up \cite{furuta1992swing, bradshaw1996swing}, and other small spatial-time-dependent PDEs \cite{kato1979korteweg}, there remains a significant gap in their application to complex structural systems. Specifically, existing approaches have not been extended to discrete spatial systems with continuous temporal behavior, which are typically governed by multiple PDEs, such as Timoshenko beam theory \cite{hutchinson2001shear} and Kirchhoff–Love shell theory \cite{hart1970linear}.

\subsection{Contributions and Paper Structure}\label{Intro_Contri}

In this work, we address the challenging problem of modeling time-dependent structural responses under dynamic loading—an area where existing machine learning solutions are currently limited. We propose a MIONet architecture specifically designed to handle the dual complexity of discrete spatial configurations and continuous temporal evolution in structural systems. Our approach introduces a unified branch network alongside two decoupled trunk networks: a spatial trunk that encodes structural geometry and a temporal trunk that captures dynamic evolution across time. This dual-trunk formulation ensures continuity in both space and time while enhancing generalization under varying moving loads and velocities. Furthermore, we move beyond conventional element-wise operations by enabling matrix-based interactions between branch and trunk outputs, allowing for richer representations and more accurate full-field predictions. Additionally, we introduce a novel physics-informed learning framework that incorporates precomputed mass, damping, and stiffness matrices to enforce dynamic equilibrium, ensuring the output accurately follows the underlying physics of the system. To further enhance computational efficiency, we employ the Schur complement to reduce problem size, making real-time dynamic structural analysis feasible. With this approach, we aim to predict structural displacements and rotations at each mesh point for every time step under varying moving loads and velocities, providing a highly efficient alternative to traditional numerical simulations.

To validate the effectiveness of the proposed framework, we test it on the same structures as those in our previous study: a two-dimensional (2D) beam structure and a three-dimensional (3D) model of a real bridge, KW-51, located in Leuven, Belgium. The organization of the paper is as follows: Section \ref{Proposed_Section} outlines the methodology,  including the MIONet architecture and the training strategies employed. Section \ref{toybridge} presents the application of the proposed method to a 2D beam structure, covering FEM modeling, data generation and processing, a comparative study of GRU-DeepONet and the proposed model, followed by the results and discussion. Section \ref{KW51_Sec} extends the approach to the real-world KW-51 structure, detailing FEM modeling, validation, data generation, data processing, training with various loss function combinations, and the results. Finally, Section \ref{conclu} presents the conclusions of the study.

\section{Proposed Method}\label{Proposed_Section}

This section outlines the methodology adopted in this study, detailing the overall approach for predicting structural dynamic responses, the proposed MIONet, and the various training strategies based on loss functions and domain size.

\subsection{Overall Methodology}\label{Overall_Method}

\begin{figure}[!htb]
    \centering
    \includegraphics[width=0.95\textwidth]{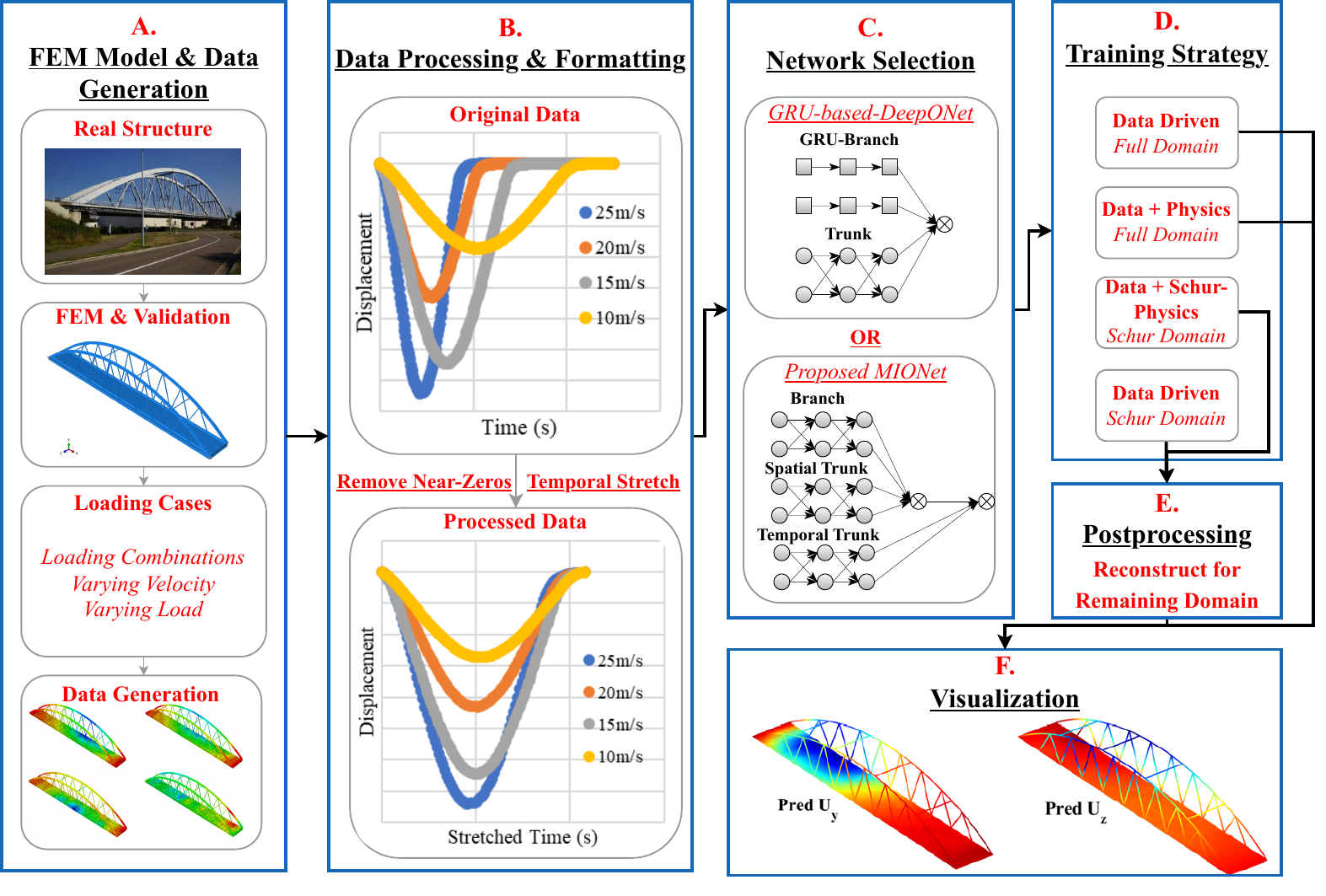}
    \caption{Flowchart illustrating the proposed methodology} 
    \label{flowchart}
\end{figure}

The proposed framework follows these steps (Figure \ref{flowchart}):

\begin{enumerate}
    \item \textbf{FEM Model Validation and Data Generation:}
    \begin{itemize}
        \item Use a previously validated FEM model \cite{ahmed2025physics, qiu2024damage}.
        \item Generate additional data under varied dynamic loading scenarios (velocities, magnitudes, configurations) (Figure \ref{flowchart}\textbf{A}).
    \end{itemize}
    
    \item \textbf{Data Processing and Temporal Alignment:}
    \begin{itemize}
        \item Remove near-zero response tails from time histories corresponding to different velocities.
        \item Apply a temporal stretching scheme to align all sequences with a common reference duration (Figure \ref{flowchart}\textbf{B}).
        \item Introduce a velocity-dependent scaling factor \(\lambda\) to preserve physical consistency across samples.
        \item Resample all sequences to a uniform time step, ensuring consistent temporal input for the model.
    \end{itemize}

    \item \textbf{Input-Output Data Formatting:}
    \begin{itemize}
        \item Map scalar inputs (load, velocity), spatial mesh discretization, and temporal discretization to a 4D input matrix.
        \item Define outputs as six structural response variables (displacements and rotations) at each mesh point and time step.
    \end{itemize}
    
    \item \textbf{Network Selection and Optimization:}
    \begin{itemize}
        \item Choose a suitable operator learning model (Figure \ref{flowchart}\textbf{C}):
        \begin{itemize}
            \item GRU-based DeepONet
            \item Proposed MIONet
        \end{itemize}
        \item Conduct a detailed parametric study to optimize architecture and hyperparameters.
    \end{itemize}
    
    \item \textbf{Training Strategies:}
    \begin{itemize}
        \item \textbf{Data-Driven (Full Domain):} Minimizes data-based MSE loss across the entire spatial domain using available response data.
        \item \textbf{Data-Driven + Physics-Informed (Full Domain):} Combines data-based MSE loss with a physics-based loss by enforcing dynamic equilibrium throughout the full domain.
        \item \textbf{Data-Driven + Physics-Informed (Schur Domain):} Applies both data and physics-based losses over a reduced Schur domain during training; the full solution is reconstructed in post-processing by satisfying the equilibrium equations.
        \item \textbf{Data-Driven (Schur Domain):} Trains solely on data-based MSE loss over the Schur domain, with full-domain responses reconstructed in post-processing using the governing physics.
    \end{itemize}
    
    \item \textbf{Post-Processing and Visualization:}
    \begin{itemize}
        \item Reconstruct responses over the entire domain from reduced Schur domain predictions (Figure \ref{flowchart}\textbf{E}).
        \item Visualize time-dependent structural behavior using scientific tools (e.g., 3D plots, ODB file rewriting) (Figure \ref{flowchart}\textbf{F}).
    \end{itemize}
\end{enumerate}

\subsection{Multiple-Input Operator (MIONet)}\label{DEEPONET}

In 2021, Lu et al. \citep{lu2021learning} introduced DeepONet, a pioneering operator learning architecture inspired by the universal approximation theorem for operators \citep{chen1995universal, back2002universal}. DeepONet provides a straightforward and intuitive model architecture that trains efficiently and offers a continuous representation of target output functions, independent of resolution. The typical architecture consists of two sub-networks: a branch network that encodes the input function \( u(x_i) \) at fixed points \( x_i, i = 1, \dots, m \), and a trunk network that encodes the locations \( a \) for the real-value output function \( G(u)(a) \), where \( G \) is the operator acting on the input function \( u \), producing the output function \( G(u) \).

The DeepONet is designed to learn operators from a single Banach space, where the operator's input is a single function. While some recent work \cite{kovachki2023neural} allows the input function to be a vector-valued function, it still requires that all components of the input function be defined on the same domain. This limitation restricts the types of operators that can be learned. For example, when solving PDEs, operators mapping both the initial condition and boundary condition to the PDE solution cannot be easily learned, as the initial and boundary conditions are defined on separate domains—initial conditions on one domain and boundary conditions on another.

To overcome this limitation, Jin et al. \cite{jin2022mionet} proposed MIONet, which enables input functions to be defined on multiple Banach spaces, meaning that multiple branch networks can be used to encode different components of the input function. The output, however, is defined on a single Banach space. The primary distinction between DeepONet and MIONet is that DeepONet uses a single model for both input and output spaces, whereas MIONet splits the input and output spaces, using separate models for each. In this approach, each input has a corresponding model, and for the outputs, there can be either one or multiple models. These models are then combined to compute the output. This enables the model to capture intricate relationships between various system components (e.g., initial conditions and boundary conditions). Specifically, in the case of PDE systems, the initial condition is encoded in the branch network, while the boundary conditions—spatial and temporal—are encoded in two separate trunks. The results in MIONet are obtained via element-wise multiplication, also known as the Hadamard product, between the outputs of the branch network and the two trunks.

\begin{figure}[!htb]
    \centering
    \includegraphics[width=0.6\textwidth]{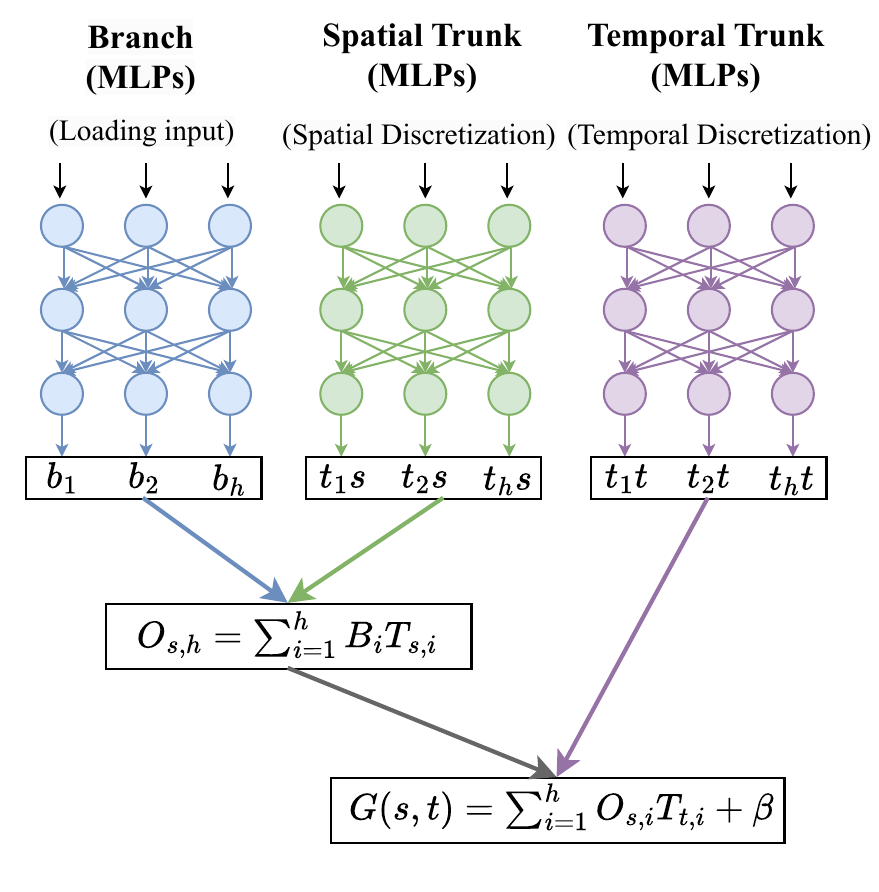}
    \caption{Architecture of the proposed  MIONet, incorporating two trunk networks and a single branch network based on MLPs}
    \label{deeponet_fig}
\end{figure}

Based on this concept, we propose a MIONet that encodes all loading conditions through a single branch network, while simultaneously decoupling the output domain into two distinct representations: spatial and temporal. In our formulation, the branch network \( [B] \) encodes the input loading conditions—comprising velocity and load values—into a latent representation \( [b_1, b_2, \ldots, b_h] \). The first trunk network \( [T_s] \), responsible for spatial encoding, maps the spatial coordinates \( s \) to a feature vector \( [t_1^{(s)}, t_2^{(s)}, \ldots, t_h^{(s)}] \), while the second trunk network \( [T_t] \) maps temporal inputs \( t \) to the temporal embedding \( [t_1^{(t)}, t_2^{(t)}, \ldots, t_h^{(t)}] \). The final output is computed through a stepwise interaction controlled via an Einsum-based approach. This approach first handles interactions between the branch network and the spatial trunk, and then incorporates the temporal trunk, followed by the addition of a bias term (Figure \ref{deeponet_fig}). The mathematical formulation is as follows:

\begin{equation}
    G(s, t) = \sum_h \sum_h \left( B_h \cdot T_s^h \right) \cdot T_t^h + \beta
\end{equation}

Where \( B_h \) is the branch output, representing the feature embedding obtained from the branch network. \( T_s^h \) is the spatial trunk output, which generates the feature embedding based on the spatial coordinates \( s \). \( T_t^h \) is the temporal trunk output, which produces the feature embedding based on the temporal coordinate \( t \). The hidden dimension \( h \) is summed over twice, capturing the interactions between the branch output and both spatial and temporal trunks.

This approach offers several key benefits. First, it allows the model to learn structured dependencies between spatial and temporal domains separately, ensuring that contributions from both domains are fully captured. Additionally, by capturing spatial features first, the model ensures that spatial information is thoroughly processed before introducing temporal dynamics, which is crucial for accurate representation in dynamic simulations. This separation also facilitates a more straightforward interpretation of the model's behavior, as spatial and temporal contributions can be analyzed independently. Finally, the approach provides flexibility for future extensions, allowing for the incorporation of additional layers or techniques, such as Fourier layers, attention mechanisms, or wavelet transforms, to further enhance the model's performance.

\subsection{Training Strategies}\label{New_loss_Section}
In this work, we propose four distinct training strategies, each defined by the type of loss function and the spatial domain over which it is applied (Table \ref{tab:training_strategies}). The choice of loss function is critical to the performance of the machine learning model, as it governs the optimization process by influencing how the network adjusts its weights during backpropagation.

We consider a combination of data-driven and physics-informed loss functions, with weights applied, either over the full spatial domain or a reduced Schur domain. The total loss function is formulated as:

\begin{equation}
\label{Loss}
 \mathcal{L}_{\text{Total}} = w_1\mathcal{L}_{\text{data}} + w_2\mathcal{L}_{\text{physics}}\\
\end{equation}

We propose novel structural matrix-based physics-informed loss functions to handle dynamic loading, which is a fundamental expansion of the static loading loss function previously presented by the authors in \cite{ahmed2025physics}. This expansion is achieved by incorporating pre-calculated mass, damping, and stiffness matrices, ensuring the system satisfies dynamic equilibrium without relying on PDE-based loss functions.

\begin{table}[]
\centering
\caption{Summary of Training Strategies}
\label{tab:training_strategies}
\begin{tabular}{|c|c|c|c|c|}
\hline
\textbf{Strategy}                                       & \textbf{Description}                                                                                         & \textbf{Domain} & \textbf{\begin{tabular}[c]{@{}c@{}}Loss \\ Function\end{tabular}} & \textbf{\begin{tabular}[c]{@{}c@{}}Post-\\ processing\end{tabular}} \\ \hline
\begin{tabular}[c]{@{}c@{}}DD\\ (Full)\end{tabular}     & \begin{tabular}[c]{@{}c@{}}Purely data-driven \\ using the full \\ spatial domain\end{tabular}               & Full            & \begin{tabular}[c]{@{}c@{}}MSE\\ (DD)\end{tabular}                & -                                                                   \\ \hline
\begin{tabular}[c]{@{}c@{}}DD+PI\\ (Full)\end{tabular}  & \begin{tabular}[c]{@{}c@{}}Combines data and\\ physics loss over \\ full domain\end{tabular}                 & Full            & \begin{tabular}[c]{@{}c@{}}MSE\\ (DD+PI)\end{tabular}             & -                                                                   \\ \hline
\begin{tabular}[c]{@{}c@{}}DD+PI\\ (Schur)\end{tabular} & \begin{tabular}[c]{@{}c@{}}Trains on the Schur\\  domain with combined \\ data and physics loss\end{tabular} & Schur           & \begin{tabular}[c]{@{}c@{}}MSE\\ (DD+PI)\end{tabular}             & Yes                                                                 \\ \hline
\begin{tabular}[c]{@{}c@{}}DD\\ (Schur)\end{tabular}    & \begin{tabular}[c]{@{}c@{}}Data-driven only \\ on Schur nodes, physics \\ used post-training\end{tabular}    & Schur           & \begin{tabular}[c]{@{}c@{}}MSE\\ (DD)\end{tabular}                & Yes                                                               \\ \hline
\end{tabular}
\end{table}

\subsubsection{Data-Driven (Full Domain)}\label{DD_Loss}
The Data-Driven Full Domain (DD-Full) training strategy is based purely on minimizing the error between the predicted and true responses over the entire spatial domain. 

Data-driven loss functions are commonly used to ensure that the ML network learns based on the provided data \cite{ahmed2022seismic,xu2020prediction,lei2019fault,xu2019automatic}. These functions ensure that the ML network's training adheres strictly to the available data, with the quality of predictions being dependent on the data used during training. In this work, the DD-Full approach utilizes the Mean Squared Error (MSE) loss, defined as:

\begin{equation}
\label{Loss_data}
\mathcal{L}_{\text{DD-Full}} = \frac{1}{N} \sum_{i=1}^N \left( \|G_{\text{True}}(u_i)(s,t) - G(u_i)(s,t)\|^2 \right)
\end{equation}

Here, \( G(u_i)(s,t) \) represents the predicted output function at point \( s \) and time \( t \), while \( G_{\text{True}}(u_i)(s,t) \) is the true function value (from FEM) at the same point \( s \) and time \( t \). The parameter \( N \) denotes the total number of functions processed by the branch network. This loss function aims to minimize the squared error between predicted values and actual values, ensuring both spatial and temporal accuracy.

While this DD-Full approach ensures the network learns from the data, it is also crucial to incorporate the underlying physical principles of the system to produce realistic predictions. Purely data-driven loss functions may sometimes fail to ensure that predictions adhere to the governing physics, especially when the training data is of insufficient quality or quantity. Moreover, for large-scale structural systems, data-driven learning can become computationally prohibitive due to the high dimensionality of the problem—requiring predictions for thousands of degrees of freedom (DOFs) at each time step, which significantly increases both data volume and model complexity. To overcome these limitations, we introduce three learning strategies based on physics-informed loss functions.

\subsubsection{Data-Driven + Physics-Informed (Full Domain)}\label{PI_Loss}

The second strategy, Data-Driven + Physics-Informed Full Domain (DD+PI-Full), integrates physical principles directly into the training process to ensure that predictions not only fit the available data but also obey the governing laws of structural dynamics across the entire spatial domain.

Researchers have integrated physics into DD approaches, leading to PINNs \cite{raissi2019physics, abueidda2021meshless}, ensuring ML predictions align with underlying physics rather than just training data. Typically, physics is incorporated by formulating loss as PDE residuals at collocation points \cite{niaki2021physics, henkes2022physics, rao2021physics} or using variational and energy methods \cite{samaniego2020energy, nguyen2020deep, abueidda2022deep}. However, extending these approaches to real-world structures composed of discrete members (e.g., beams and shells) introduces several challenges. First, defining governing PDEs across non-continuous domains is non-trivial. Second, energy-based or variational formulations become computationally prohibitive due to large degrees of freedom and structural complexity. Finally, incorporating time-dependent behavior, particularly inertia and damping, into the loss function adds further dimensional and numerical complexity. To address these issues, we introduce a new physics-informed loss formulation that builds on the foundational idea of using precomputed structural matrices \cite{ahmed2025physics}. Unlike the static case, our dynamic formulation incorporates mass, damping, and stiffness matrices directly into the loss function, enabling enforcement of the full dynamic equilibrium equation across all time steps. 

The dynamic equilibrium equation is fundamental in structural dynamics, governing the relationship between inertia, damping, and stiffness to ensure physically consistent system responses. Derived from Newton’s Second Law, the equation is expressed as:  

\begin{equation}\label{Orignal_Dyn_Eq}
    Ma(t) + Cv(t) + K u(t) = F(t)  
\end{equation}

where \( M \) is the mass matrix representing inertia effects, \( C \) is the damping matrix accounting for energy dissipation, and \( K \) is the stiffness matrix governing elastic restoring forces. The displacement response is given by \( u(t) \), with its first and second derivatives, \( v(t) \) and \( a(t) \), representing velocity and acceleration, respectively. \( F(t) \) denotes the external force vector, such as applied loads or excitations. This equation ensures that, at any time \( t \), the total external force is balanced by the sum of inertial, damping, and elastic forces, maintaining dynamic equilibrium.

Since the objective of this study is to predict displacements and rotations using ML model, the output primarily comprises displacement responses \( u(t) \). To ensure that the solution satisfies the dynamic equilibrium equation, velocity \( v(t) \) and acceleration \( a(t) \) must be calculated. Various numerical schemes exist for this purpose, including the Newmark method and the Hilber-Hughes-Taylor (HHT) method. In this study, the HHT method is used within the Abaqus finite element simulations to generate the dynamic response data, as it introduces numerical damping that helps mitigate high-frequency oscillations in implicit dynamic analyses.

The HHT method extends the classical Newmark method by incorporating numerical damping and stability adjustments, modifying the standard dynamic equilibrium equation (Eq. \ref{Orignal_Dyn_Eq}). While traditional methods directly solve for acceleration, velocity, and displacement relationships, the HHT method employs an implicit formulation that includes damping factors for improved accuracy and stability in dynamic problems. The modified dynamic equilibrium equation is given as:

\begin{equation}
\label{Modified_Dyn_Eq}
    M a_{t+1} + (1+\alpha) C v_{t+1} + (1+\alpha) K u_{t+1} = F_{t+1} + \alpha (F_t - C v_t - K u_t)  
\end{equation}

where \( \alpha \) is a numerical damping parameter, typically chosen within the range \( -\frac{1}{3} \leq \alpha \leq 0 \). Higher values of \( \alpha \) increase damping, thereby enhancing stability in systems with high-frequency oscillations or stiffness; however, this comes at the cost of reduced accuracy. When \( \alpha = 0 \), the method reverts to the classical Newmark formulation, and the modified dynamic equilibrium form (Eq. \ref{Modified_Dyn_Eq}) simplifies back to its original form (Eq. \ref{Orignal_Dyn_Eq}).

In the HHT method, the acceleration \( a_{t+1} \) and velocity \( v_{t+1} \) at the next time step depend on their respective values from the current step \( t \), along with the displacement at both the current \( u_{t} \) and next \( u_{t+1} \) time steps. These relationships are given as:

\begin{equation}
\label{Acc_Eq}
    a_{t+1} = \frac{1}{\beta \Delta t^2} \left( u_{t+1} - u_t - \Delta t v_t \right) - \frac{(1 - 2\beta)}{2\beta} a_t
\end{equation}

\begin{equation}
\label{Velo_Eq}
    v_{t+1} = v_t + \Delta t \left( (1-\gamma) a_t + \gamma a_{t+1} \right) 
\end{equation}

The HHT method introduces numerical damping through three key parameters: \( \alpha \), \( \beta \), and \( \gamma \), which control the stability and accuracy of the integration scheme. These parameters are defined as \( \beta = 0.25 (1 - \alpha)^2 \) and \( \gamma = 0.5 - \alpha \). The parameter \( \beta \) influences how acceleration is weighted in the integration process, while \( \gamma \) governs the contribution of current and future accelerations when updating velocity. By adjusting these values, the HHT method offers a balance between accuracy and numerical stability.

To construct a loss function based on the dynamic equilibrium equation, we reformulate it in terms of effective stiffness (\( K_{\text{eff}} \)) and effective external forces (\( F_{\text{eff}} \)). This approach simplifies the standard dynamic equation by incorporating the effects of mass, damping, and stiffness into a single effective stiffness matrix. Similarly, the external force term is modified to include contributions from previous time steps, accounting for velocity, acceleration, and displacement history.

By substituting the expressions for acceleration (Eq. \ref{Acc_Eq}) and velocity (Eq. \ref{Velo_Eq}) into the dynamic equilibrium equation (Eq. \ref{Modified_Dyn_Eq}) and rearranging, we obtain the final form:

\begin{equation}
\label{Keff_Eq}
    K_{\text{eff}} u = F_{\text{eff}}
\end{equation}

where \( K_{\text{eff}} \) represents the effective stiffness matrix, which combines the contributions of mass, damping, and stiffness, while \( F_{\text{eff}} \) is the effective external force that depends on the applied load, velocity, acceleration, displacement, HHT parameters, and the time step increment.

The dynamic equilibrium equation (Eq. \ref{Keff_Eq}) assumes constant velocity and a fixed time step \( \Delta t \). However, in our problem, the speed of the moving load varies across samples, resulting in different total simulation durations. To enable uniform input to the temporal trunk of MIONet, we discretize the temporal domain using a fixed number of time steps, regardless of the actual simulation duration. This introduces a discrepancy: each step now corresponds to a different physical time duration, thus invalidating the original assumptions of constant \( \Delta t \).

To resolve this, we introduce a velocity-dependent scaling factor \( \lambda_v \), defined as the ratio between a reference speed and the actual speed. This factor allows us to stretch or compress the response in time so that each simulation aligns to a common reference time domain. This ensures that the dynamic quantities remain consistent and physically meaningful across varying velocities.

Based on the \( \lambda_v \) factor the modified acceleration and velocity can be given as:

\begin{equation}
\label{Modi_Acc_Eq}
a_{t+1} = \frac{\lambda_v^2}{\beta \Delta t^2} \left( u_{t+1} - u_t - \frac{\Delta t}{\lambda_v} v_t \right) - \frac{(1 - 2\beta)}{2\beta} a_t 
\end{equation}

\begin{equation}
\label{Modi_Velo_Eq}
v_{t+1} = v_t + \frac{\Delta t}{\lambda_v} \left( (1 - \gamma) a_t + \gamma a_{t+1} \right)
\end{equation}

The effective stiffness matrix is:

\begin{equation}
\label{Modi_Keff_Explain_Eq}
K_{\text{eff}} = \frac{\lambda_v^2 M}{\beta \Delta t^2} + (1 + \alpha) \frac{\lambda_v C \gamma}{\beta \Delta t} + (1 + \alpha) K
\end{equation}

The effective force term is:

\begin{equation}
\label{Modi_Feff_Explain_Eq}
\begin{aligned}
    F_{\text{eff}} &= F_{t+1} + \alpha(F_t - K u_t) + M \left(\frac{\lambda_v^2 u_t}{\beta \Delta t^2} + \frac{\lambda_v v_t}{\beta \Delta t} + \frac{(1-2\beta) a_t}{2\beta} \right) \\
    &+ C \left[ v_t \left( \frac{\alpha \gamma}{\beta} + \frac{\gamma}{\beta} - 2\alpha - 1 \right) + a_t \frac{\Delta t}{\lambda_v} \left( \frac{\alpha \gamma}{2\beta} + \frac{\gamma}{2\beta} - \alpha - 1 \right) + \frac{\lambda_v u_t}{\Delta t} \left( \frac{\gamma}{\beta} + \frac{\alpha \gamma}{\beta} \right) \right]
\end{aligned}  
\end{equation}

By introducing \( \lambda_v \), we ensure that the dynamic equilibrium equations hold across different moving speeds, allowing MIONet to process all cases under a unified temporal discretization.

The loss function for this strategy is designed to enforce the dynamic equilibrium equation (Eq.~\ref{Keff_Eq}) by minimizing MSE between both sides of the equation. Incorporating both the data and physics components, the final loss function for the DD+PI-Full strategy becomes:

\begin{equation}
\label{DD+PI_FullLoss}
\begin{aligned}
\mathcal{L}_{\text{DD+PI-Full}} &= 
w_1 \left( \frac{1}{N} \sum_{i=1}^N \left\| G_{\text{True}}(u_i)(s,t) - G(u_i)(s,t) \right\|^2 \right) \\
&+ w_2 \left( \frac{1}{N} \sum_{i=1}^N \left\| K_{\text{eff}} G(u_i)(s,t) - F_{\text{eff}} \right\|^2 \right)
\end{aligned}
\end{equation}

This formulation ensures that the network predictions align with both the available data and the physical laws of dynamic equilibrium across the full spatial domain. However, enforcing equilibrium at all DOFs necessitates large-scale matrix operations, often involving thousands of DOFs per time step. This makes training computationally intensive and memory demanding. To mitigate this, we introduce a Schur complement-based approach that reduces the dimensionality of the problem while preserving physical consistency, thereby improving efficiency without sacrificing accuracy.

\subsubsection{Data-Driven + Schur Complement Approach (Schur Domain)}\label{SchurTrain}

The Schur complement \cite{CARLSON1986257} is a widely used technique in linear algebra that facilitates the reduction of large systems of equations, making them computationally more efficient to solve. It plays a crucial role in domain decomposition methods, such as the Finite Element Tearing and Interconnecting (FETI) method, where it enables independent solutions of subdomains while ensuring consistency at interfaces \cite{langer2005coupled, pechstein2012finite}. Similarly, in the Boundary Element Method (BEM), the Schur complement is utilized to simplify boundary integral equations by reformulating them in block matrix form \cite{mobasher2016adaptive,hackbusch2005direct}.  

In our previous work, this approach was successfully applied to static loading conditions \cite{ahmed2025physics}. Here, we extend its application to dynamic systems using the simplified dynamic equilibrium equation (Eq. \ref{Keff_Eq}). The objective is to reduce the training domain of the problem while preserving the essential characteristics of the system dynamics.  

To achieve this, we partition the system into a block matrix form:  

\begin{equation}
 \label{Matrix}
\begin{bmatrix} 
K_{[II]} & K_{[IN]} \\ 
K_{[NI]} & K_{[NN]} 
\end{bmatrix}_{\text{eff}}
\begin{bmatrix} 
u_{[I]} \\ 
u_{[N]} 
\end{bmatrix}%_{n+1}
=
\begin{bmatrix} 
F_{[I]} \\ 
F_{[N]} 
\end{bmatrix}_{\text{eff}}
\end{equation}

where \( u_{[I]} \) represents the DOFs of interest for which DeepONet is trained to predict results, while \( u_{[N]} \) corresponds to the remaining DOFs, which can be reconstructed through post-processing. By expanding and reorganizing the system, we obtain the reduced form:  

\begin{equation}
\label{Main_Eq}
S_{\text{eff}} u_{[I]} = F_{[C], \text{eff}}
\end{equation}

where the Schur complement \(S_{\text{eff}}\) of the system is given by:  

\begin{equation}
\label{S_expand}
S_{\text{eff}} = K_{[II] \text{eff}} - K_{[IN] \text{eff}} K_{[NN] \text{eff}}^{-1} K_{[NI] \text{eff}}
\end{equation}

and the modified force term \(F_{[C] \text{eff}}\) is defined as:  

\begin{equation}
\label{Fc_expand}
F_{[C] \text{eff}} = F_{[I] \text{eff}} - K_{[IN] \text{eff}} K_{[NN] \text{eff}}^{-1} F_{[N] \text{eff}}
\end{equation}

where \( F_{[I]} \) and \( F_{[N]} \) are expressed as:

\begin{equation}
\label{F_I_F_N}
\begin{aligned}
    F_{[I] \text{eff}, t+1} &= F_{[I] t+1} + \alpha \left( F_{[I] t} - K_{[II,IN]} u_{[I,N] t} \right) \\
    &+ M_{[II,IN]} \left( \frac{\lambda_v^2 u_{[I,N] t}}{\beta \Delta t^2}
    + \frac{\lambda_v v_{[I,N] t}}{\beta \Delta t} \right) \\
    &+ \frac{(1-2\beta) a_{[I,N] t}}{2\beta} + C_{[II,IN]} \left[ 
    v_{[I] t} \left(\frac{\alpha \gamma}{\beta} + \frac{\gamma}{\beta} - 2\alpha -1 \right) \right. \\
    &\quad + a_{[I,N] t} \frac{\Delta t}{\lambda_v} \left(\frac{\alpha \gamma}{2\beta} + \frac{\gamma}{2\beta} - \alpha -1 \right)
    \quad \left. + \frac{\lambda_v u_{[I,N] t}}{\Delta t} \left(\frac{\gamma}{\beta} + \frac{\alpha \gamma}{\beta} \right) \right] 
\end{aligned}
\end{equation}

\begin{equation}
\label{F_N}
\begin{aligned}
    F_{[N] \text{eff}, t+1} &= F_{[N] t+1} + \alpha \left( F_{[N] t} - K_{[IN,NN]} u_{[I,N] t} \right) \\
    &+ M_{[IN,NN]} \left( \frac{\lambda_v^2 u_{[I,N] t}}{\beta \Delta t^2} 
    + \frac{\lambda_v v_{[I,N] t}}{\beta \Delta t} \right) \\
    &+ \frac{(1-2\beta) a_{[I,N] t}}{2\beta} + C_{[IN,NN]} \left[ 
    v_{[I,N] t} \left(\frac{\alpha \gamma}{\beta} + \frac{\gamma}{\beta} - 2\alpha -1 \right) \right. \\
    &\quad + a_{[I,N] t} \frac{\Delta t}{\lambda_v} \left(\frac{\alpha \gamma}{2\beta} + \frac{\gamma}{2\beta} - \alpha -1 \right) \quad \left. + \frac{\lambda_v u_{[I,N] t}}{\Delta t} \left(\frac{\gamma}{\beta} + \frac{\alpha \gamma}{\beta} \right) \right] 
\end{aligned}
\end{equation}

The post-processing solution for the remaining degrees of freedom is given by:

\begin{equation}
\label{u_N}
\begin{aligned}
    u_{[N]} &= K_{[NN] \text{eff}}^{-1} \left( F_{[N] \text{eff}} - K_{[NI] \text{eff}} u_{[I]} \right)
\end{aligned} 
\end{equation}

This process must be iterated across all time steps. To ensure that Eq. \ref{S_expand} is satisfied at each timestep during network training, and that the available data is learned adequately, the following expression must hold:

\begin{equation}
\label{Loss_PI_Schur}
\begin{aligned}
\mathcal{L}_{\text{DD+PI-Schur}} &= 
w_1 \left( \frac{1}{N} \sum_{i=1}^N \left\| G_{\text{True}}(u_i)(s,t) - G(u_i)(s,t) \right\|^2 \right) \\
&+ w_2 \left( \frac{1}{N} \sum_{i=1}^{N} \left\| S_{\text{eff}} G_{[I]} (u_i)(s,t) - F_{[C], \text{eff}} \right\|^2 \right)
\end{aligned}
\end{equation}

\begin{figure}[!htb]
    \centering
    \includegraphics[width=0.7\textwidth]{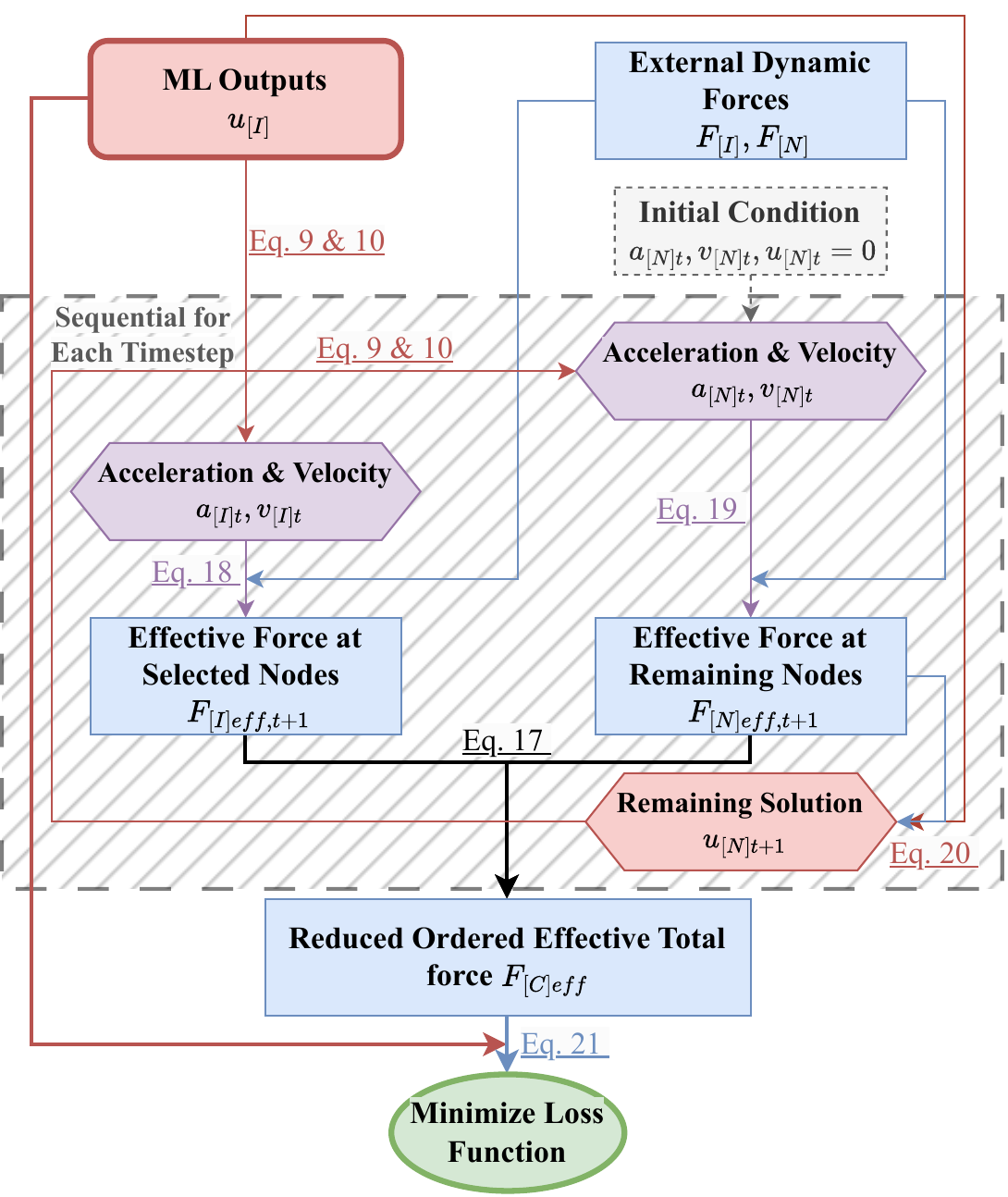}
    \caption{Flowchart for integrating the Schur-based loss function in the network training process} 
    \label{Schur_FlowChart}
\end{figure}

This loss function ensures that the prediction adheres to the available data for the Schur domain and enforces the dynamic equilibrium equation within that domain. However, obtaining \( F_{C, \text{eff}} \) requires multiple steps, as outlined in the flowchart (Figure \ref{Schur_FlowChart}).

The DD+PI-Schur training strategy proceeds in multiple sequential steps. During training, the network predicts the displacement \(u_{[I]}\), which is then used in Eqs. \ref{Modi_Acc_Eq} and \ref{Modi_Velo_Eq} to compute the acceleration \( a_{[I]} \) and velocity \( v_{[I]} \) at time \( t \). For the initial step (\( t = 0 \)), the displacement \( u_{[N]} \), acceleration \( a_{[N]} \), and velocity \( v_{[N]} \) are initialized to zero. Using the computed acceleration and velocity at each time step \( t \), the reduced effective forces \( F_{[I]\text{eff},t+1} \) and \( F_{[N]\text{eff}.t+1} \) are determined via Eqs. \ref{F_I_F_N} and \ref{F_N}.  

Once \( F_{[I]eff} \) and \( F_{[N]eff} \) at \( t+1 \) are obtained, the results are postprocessed during training to compute \( u_{[N]} \) at \( t+1 \) using Eq. \ref{u_N}, followed by an update of \( a_{[N]} \) and \( v_{[N]}\) for the \( t+1 \). This process is repeated sequentially until the final time step is reached. Ultimately, the reduced-order effective force is computed using Eq. \ref{Fc_expand}, and the Schur-based loss function (Eq. \ref{Loss_PI_Schur}) is enforced.  

While this approach is computationally more demanding than the previously derived physics-informed loss function (Eq. \ref{DD+PI_FullLoss}), it offers a significant advantage in scenarios where only sparse data is available. Even with data limited to a small subset of nodes, this method ensures that the underlying dynamic physics is satisfied across the entire domain. This is particularly beneficial when the solution is known with confidence at only specific locations within the domain. Instead of relying on extensive data coverage, the network can be trained using a minimal set of domain points while still enforcing the physical constraints.  

\begin{figure}[!htb]
    \centering
    \includegraphics[width=0.8\textwidth]{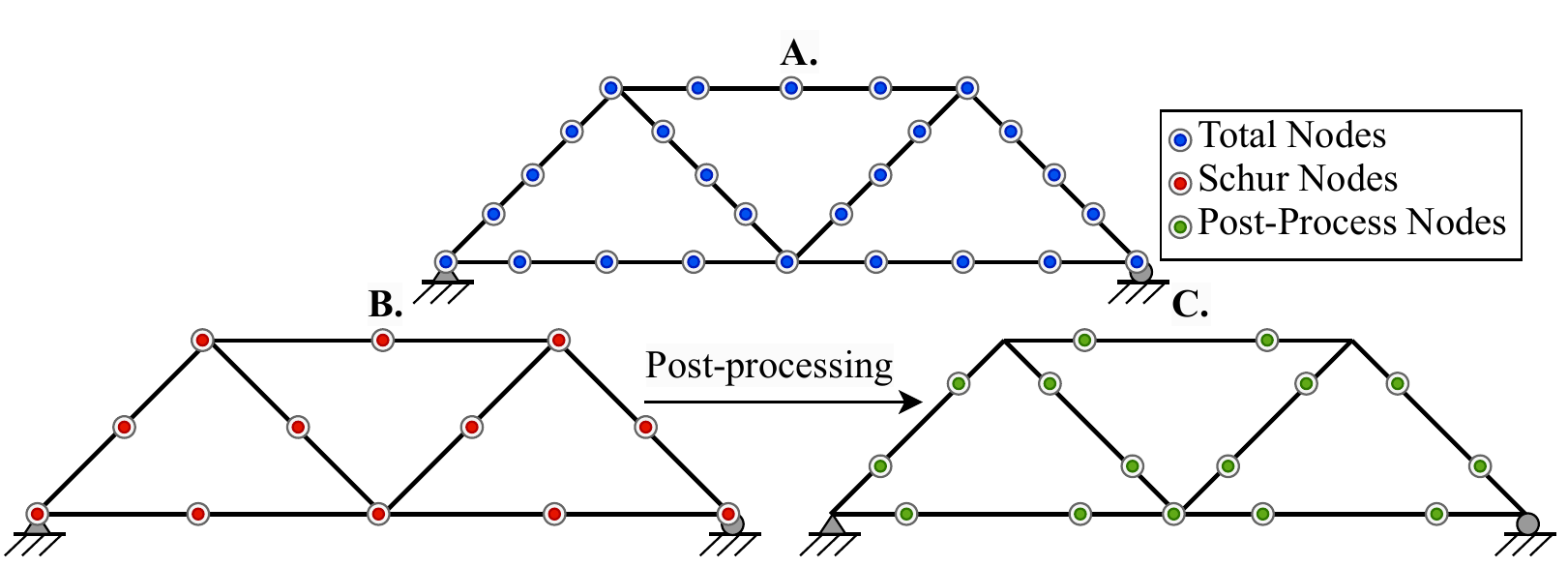}
    \caption{Illustration of the Schur Complement to reduce the size of the system: \textbf{A.} Total nodes in the structural domain, \textbf{B.} Picked nodes by applying Schur complement, \textbf{C.} The remaining nodes solution obtained by using post-processing (Eq. \ref{u_N})}
    \label{Schur_nodes}
\end{figure}

\subsubsection{Data-Driven (Schur Domain)}\label{Combi_Loss}
Data-Driven Schur Domain (DD-Schur) strategy combines pure data-driven training with the application of physics in the post-processing phase to ensure that the entire system adheres to the underlying physics. As discussed earlier, the DD+PI-Schur approach is computationally more demanding than the full-domain physics incorporation (DD+PI-Full). To optimize the efficiency of DD+PI-Schur, we modified the approach by initially training the network using a purely data-driven loss for the Schur nodes (Figure \ref{Schur_nodes}). This is highly efficient, as the spatial training domain is very small, making the training process much easier. Once the network is trained, the solution for the remaining nodes can be reconstructed using Eq. \ref{u_N}, following the procedure outlined in Figure \ref{Schur_FlowChart}.

For this training strategy, the loss function is identical to the data-driven MSE loss function, but it is restricted to the Schur domain. This is given by:

\begin{equation}
\label{DD-Schur_loss_Fun}
\mathcal{L}_{\text{DD-Schur}} = \frac{1}{N} \sum_{i=1}^N \left( \|G_{\text{True}}(u_i)(s,t) - G(u_i)(s,t)\|^2 \right)
\end{equation}

Based on the reduced domain training, the final prediction is limited to the selected nodes within the Schur domain. To reconstruct the solution for the entire spatial domain, it is essential to incorporate the accurate underlying physics. This can be achieved with high precision by utilizing the novel method proposed in this work, which leverages the structural mechanics matrix to develop the underlying physics. This approach can provide a more efficient and accurate means of capturing the system's behavior across the full domain.

\section{2D Beam Structure} \label{toybridge}
\subsection{FEM Model and Data Generation} \label{FEMmodel}

\begin{figure}[h]
    \centering
    \includegraphics[width=0.6\textwidth]{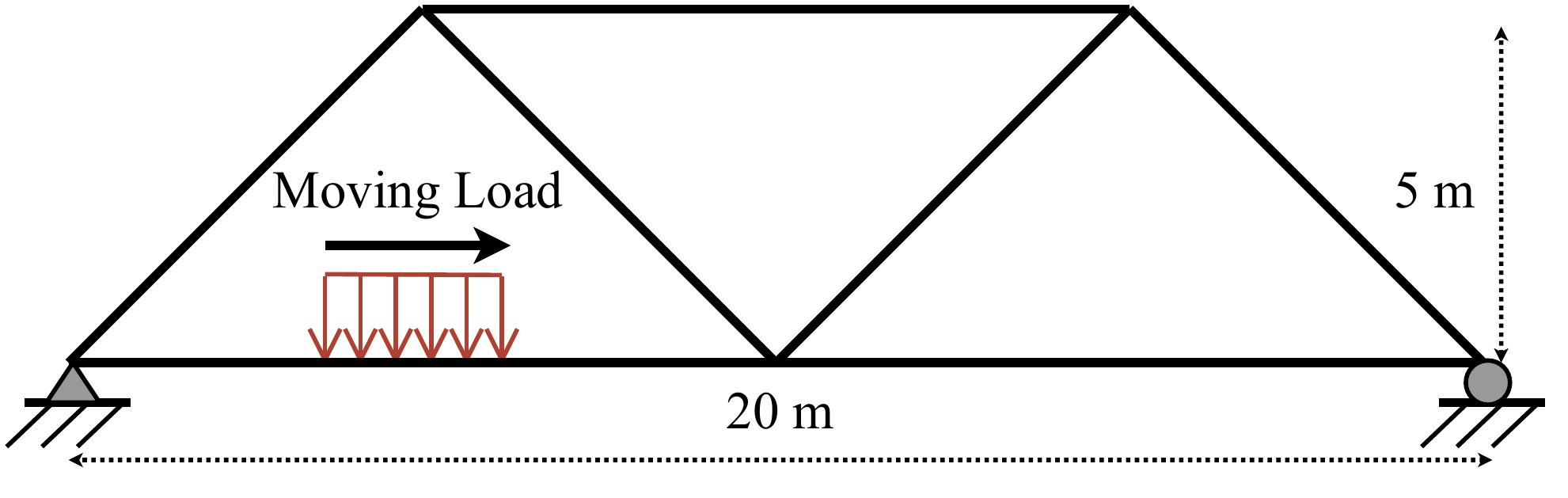}
    \caption{Schematic representation of the 2D beam structure with a single moving load}
    \label{ToyBridge}
\end{figure}

To evaluate the proposed method for response prediction, a 2D beam structure is modeled in Abaqus using Timoshenko beam elements, configured to resemble a truss structure. The structure has a horizontal span of 20 \(m\) and a vertical height of 5 \(m\), with hinged and roller boundary conditions applied at the bottom left and right supports, respectively (Figure \ref{ToyBridge}). The model consists of 56 nodes, each providing displacement outputs in the \(x\) and \(y\) directions, as well as rotational displacement about the \(z\) axis (\(U_x, U_y, R_z\)).

The beam is assigned steel material properties, including a Young’s modulus (\(E\)) of 210 \(GPa\), Poisson’s ratio (\(\nu\)) of 0.3, and density (\(\rho\)) of 7850 \(kg/m^3\). Damping is introduced using Rayleigh damping coefficients (\(\alpha = 0.1\), \(\beta = 0.05\)), selected through trial and error to achieve a realistic response under moving loads. The beam cross-section is rectangular, measuring \(400\,mm \times 250\, mm\).

A dynamic moving single-wheel load is applied from left to right across the structure. Multiple loading scenarios are considered, with velocities varying between 10 \(m/s\) and 25 \(m/s\), representing realistic moving loads. For data generation, four velocities are selected: 10 \(m/s\), 15 \(m/s\), 20 \(m/s\), and 25 \(m/s\). At each velocity, the load intensity varies between 5 \(kN/m\) and 30 \(kN/m\). For each velocity, 1,000 random samples are generated with different load values, resulting in a total dataset of 4,000 samples.

To implement the moving load in Abaqus, a DLOAD subroutine is defined, and the load is applied to a specific location as the time step progresses. The moving load length is 2 \(m\), and each simulation is run for a total duration of 2.5 seconds. The response is recorded at \(\Delta t = 0.1\) seconds, yielding 251 time steps in the temporal dimension.

\subsection{Data Processing}

At higher velocities, the moving load passes over the structure more quickly, leading to long segments of near-zero response at the end of the simulation. To minimize unnecessary computations, we define a threshold (\(1 \times 10^{-6} \, \text{m}\)) for the significant variable (\(U_y\)): if three consecutive time steps exhibit a response below this threshold, the remaining response is discarded. This results in different response durations—2.26 \(s\) for \( v = 10\ m/s\), 1.60 \(s\) for \( v = 15 \ m/s\), 1.27 \(s\) for \( v = 20 \ m/s\), and 1.08 \(s\) for \( v = 25 \ m/s\)—causing an uneven temporal distribution. To correct this, we apply a temporal stretching approach, using \( v = 10 \ m/s\) as the reference and scaling the responses of other cases accordingly, introducing stretch factors \( \lambda_{10} = 1.0000 \), \( \lambda_{15} = 1.4125 \), \( \lambda_{20} = 1.7795 \), and \( \lambda_{25} = 2.0926 \). However, this creates inconsistencies in the time step (\(\Delta t\)) across samples. To resolve this, we resample the response data to match the \(\Delta t\) of the reference velocity. Additionally, we reduce the total number of time steps from 251 to 56, as the response curves are smooth and do not require high temporal resolution. This ensures a consistent input format across different velocities, which is essential for training both the GRU-based DeepONet and the proposed MIONet.

\subsection{Comparative Study between GRU-based DeepONet and Proposed MIONet}

To determine the most suitable network for further analysis, we tested both an GRU-based DeepONet and the proposed MIONet on the same dataset using DD-Full stratergy. The input consists of velocity and applied load, structured as a matrix of size \( 4000 \times 2 \) (4000 samples, with 2 input values: velocity and load magnitude). The spatial domain is represented by a \( 56 \times 2 \) matrix (56 nodes with their \( x \)- and \( y \)-coordinates), and the temporal domain consists of 56 discrete time steps. Based on this structure, the output response is stored as \( 4000 \times 56 \times 56 \times 3 \), where 4000 corresponds to the number of samples, the first 56 represents time steps, the second 56 denotes spatial nodes, and 3 represents the displacement and rotation components (\( U_x, U_y, R_z \)).  

\begin{figure}[h]
    \centering
    \includegraphics[width=0.85\textwidth]{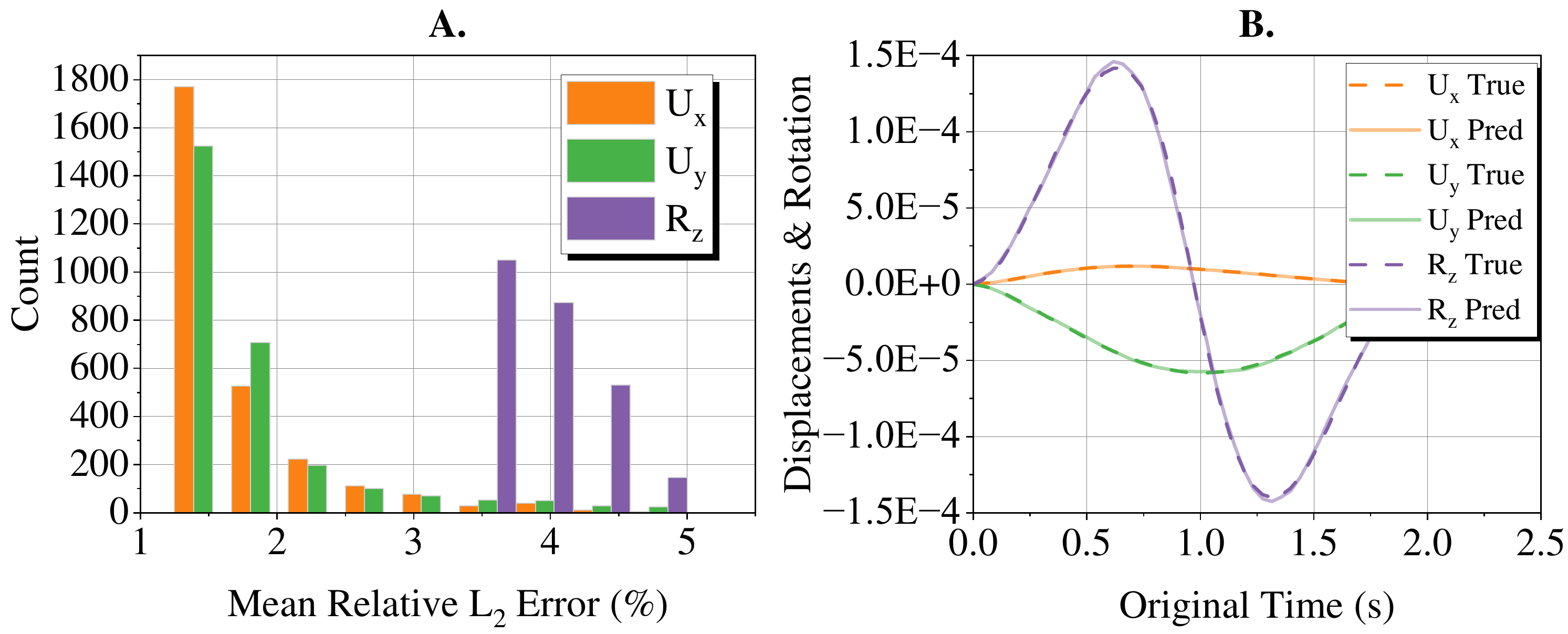}
    \caption{Predictions using the GRU-based DeepONet: \textbf{A.} Relative error histogram, \textbf{B.} Comparison of displacement and rotation at node 4 over the entire temporal domain}
    \label{GRU_DeepONET}
\end{figure}

\begin{figure}[h]
    \centering
    \includegraphics[width=0.85\textwidth]{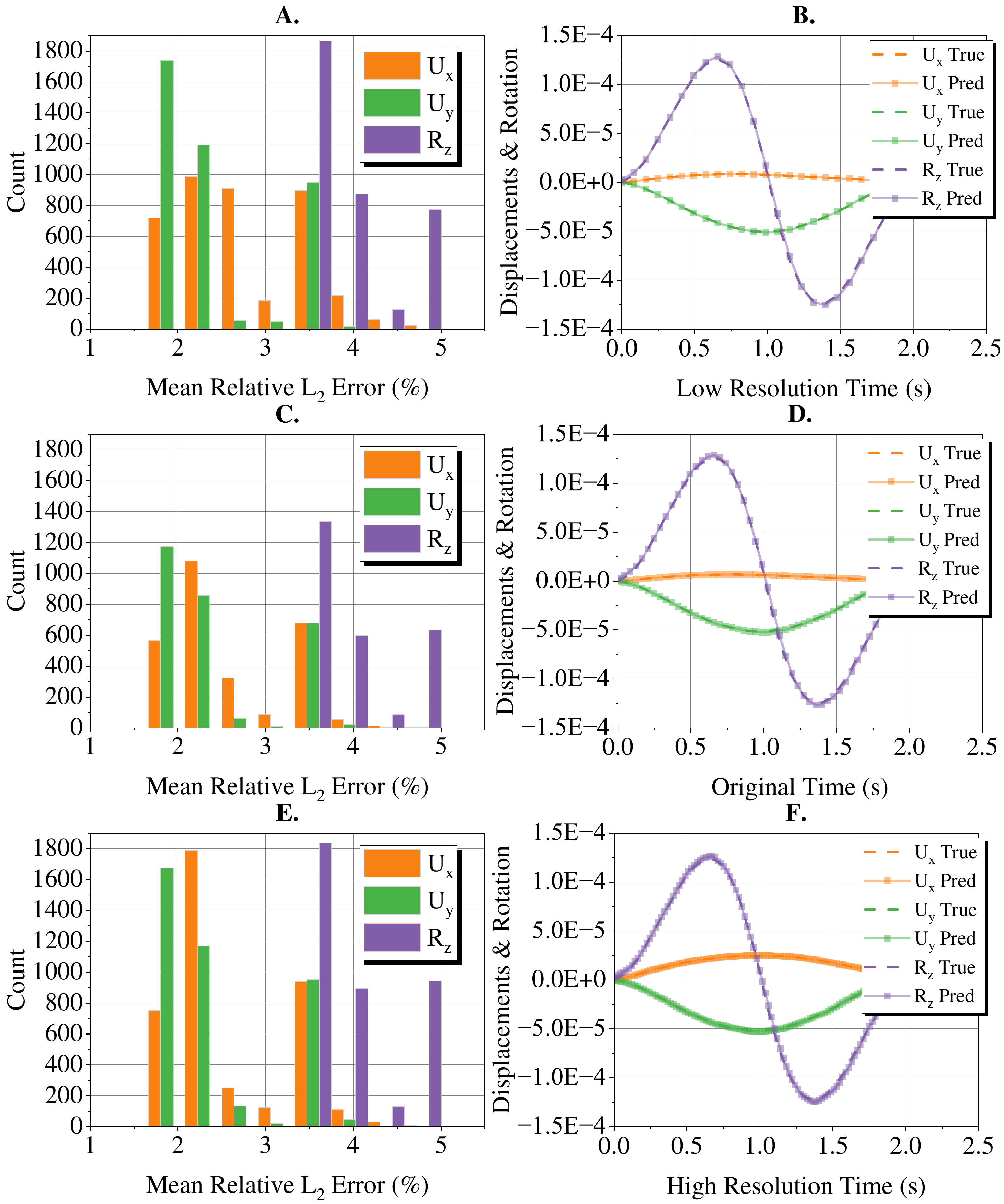}
    \caption{Predictions using the proposed MIONet (scatter points represent temporal discretization): \textbf{A.} Relative error histogram for low-resolution time steps (\( \Delta t = 0.0822 \)), \textbf{B.} Displacement and rotation comparison at node 4 in the spatial domain for low-resolution time steps (\( \Delta t = 0.0822 \)), \textbf{C.} Relative error histogram for the original simulation time steps (\( \Delta t = 0.0411 \)), \textbf{D.} Displacement and rotation comparison at node 4 in the spatial domain for the original simulation time steps (\( \Delta t = 0.0411 \)), \textbf{E.} Relative error histogram for high-resolution time steps (\( \Delta t = 0.02055 \)), \textbf{F.} Displacement and rotation comparison at node 4 in the spatial domain for high-resolution time steps (\( \Delta t = 0.02055 \))}.

    \label{MIONet-High-Original-Low}
\end{figure}

The first tested approach was the GRU-based DeepONet (\citep{he2024sequential}), which consists of a GRU branch network and a single trunk network. The GRU-based branch is responsible for capturing temporal dependencies, making it highly effective in modeling sequential loading effects. The results (Figure \ref{GRU_DeepONET}) indicate that the GRU-based DeepONet achieves slightly better pointwise accuracy compared to MIONet (Figure \ref{MIONet-High-Original-Low}), aligning with the well-known strength of RNNs in handling sequential data. However, since the GRU model processes time steps sequentially, its predictions remain fixed to predefined time steps, preventing continuous interpolation in time. Additionally, training the GRU-based DeepONet is computationally expensive, requiring 50 minutes due to the sequential nature of time step processing. In contrast, the proposed MIONet introduces a more flexible architecture with a branch network and two separate trunk networks—one for spatial dependencies and another for temporal dependencies. Unlike the GRU-based model, MIONet does not rely on sequential processing, significantly reducing computational cost. The results (Figure \ref{MIONet-High-Original-Low}) show that MIONet achieves comparable accuracy to the GRU-based DeepONet while offering a lower training time of 20 minutes. Furthermore, MIONet was tested on both low-resolution and high-resolution temporal grids (Figure \ref{MIONet-High-Original-Low}), with the low-resolution grid having a time step of \( \Delta t = 0.0822 \), the original grid having a time step of \( \Delta t = 0.0411 \), and the high-resolution grid having a time step of \( \Delta t = 0.02055 \). This results in 28 time steps for the low-resolution grid, 56 time steps for the original grid used for training, and 112 time steps for the high-resolution grid, demonstrating MIONet's ability to generate continuous outputs and interpolate at arbitrary time steps.  

While the GRU-based DeepONet offers slightly better accuracy at predefined time steps, its reliance on sequential processing and high computational cost make it less practical for dynamic structural response prediction. MIONet, on the other hand, provides a more efficient and adaptable framework by enabling continuous-time predictions without increasing computational overhead. Its ability to generalize across different temporal resolutions while maintaining a significantly lower training time makes it the preferred choice for further analysis. Therefore, despite the minor accuracy advantage of the GRU-based approach, MIONet's superior flexibility and efficiency make it the optimal solution for this study.

\subsection{Network Design and Final Configuration}  

\begin{figure}[!htb]
    \centering
    \includegraphics[width=1\textwidth]{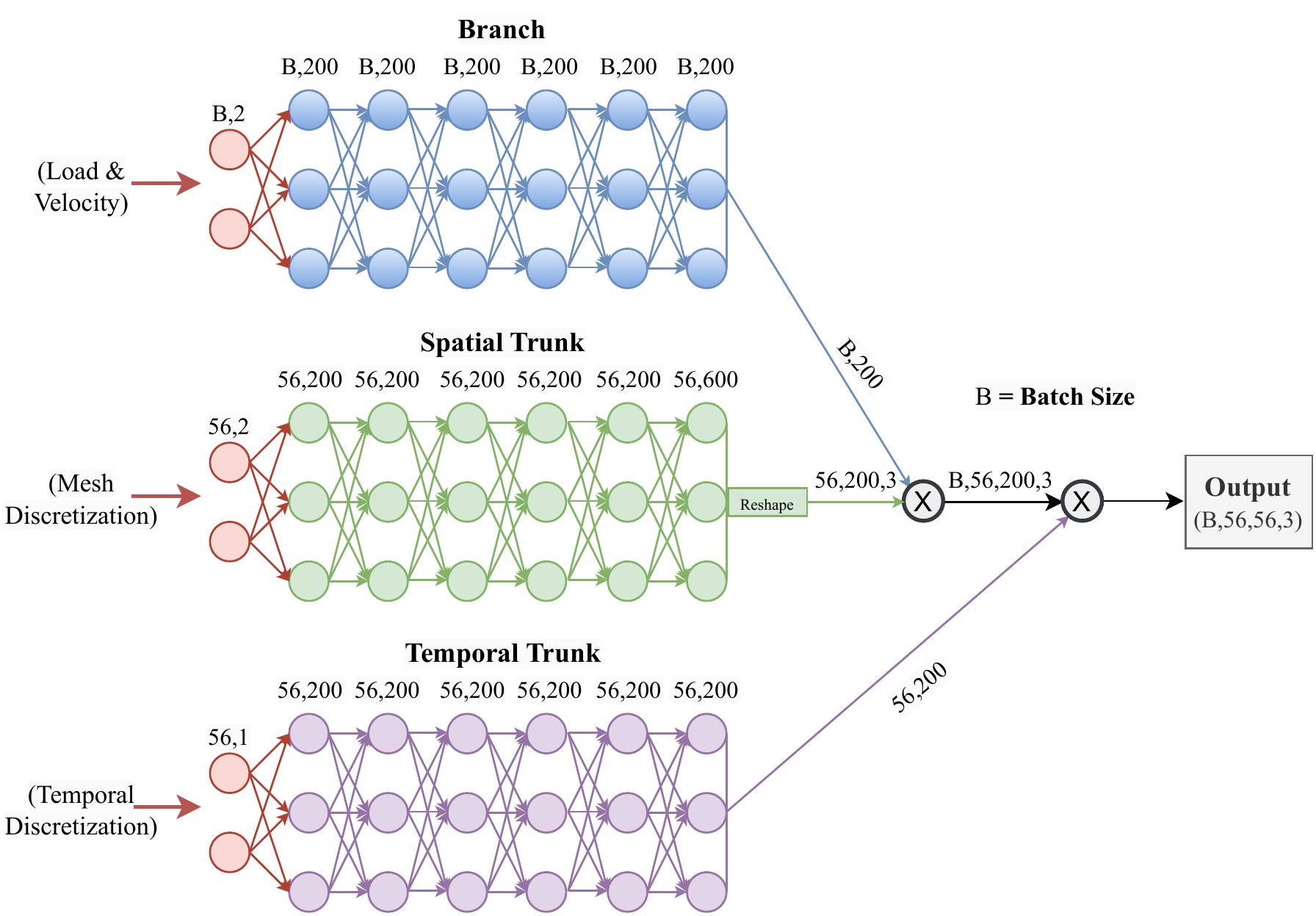}
    \caption{MIONet architecture for the 2D beam structure}
    \label{Toy_MIONET}
\end{figure}

Based on the detailed parametric study (Appendix \ref{appn1}), the final network architecture selected for training consists of three networks—one branch and two trunks—each following a simple rectangular MLP design. Each network has 6 layers with 200 neurons per layer. The optimal batch size is chosen to be 20, with a learning rate of $5 \times 10^{-4}$.  The final network configuration consists of a branch network [\textbf{2}, \textit{200, 200, 200, 200, 200, 200}], spatial trunk network [\textbf{2}, \textit{200, 200, 200, 200, 200, }[\textit{200} $\times$ \textbf{\textit{3}}]] and a temporal trunk network [\textbf{1}, \textit{200, 200, 200, 200, 200, 200}]. Here, bold values indicate the input layers, italics denote the six hidden layers, and in the spatial trunk, the bold-italic term represents the number of output functions. The final output is computed through a stepwise interaction controlled via an Einsum-based approach. The interactions are handled in two stages: first, the branch output is combined with the spatial trunk output, and then this combined representation is expanded through the temporal trunk to obtain the final prediction. Figure \ref{Toy_MIONET} illustrates the final network architecture along with the data dimensionality as it propagates through each layer. The network is trained using the ADAM optimizer, with the default initializer and the ReLU activation function. All computations were performed on the High-Performance Computing (HPC) cluster of NYUAD, Jubail, utilizing an Nvidia A100 GPU with 10 CPU cores.  

\subsection{Training and Results}

We evaluated the performance of the finalized MIONet by training it with four different configurations, each utilizing distinct data and loss function strategies: DD-Full, DD+PI-Full, DD+PI-Schur, and DD-Schur.  The DD+PI-Schur and DD-Schur strategies train on only 5 out of the 56 spatial nodes, and the responses for the remaining domain are then reconstructed through post-processing (Figure \ref{Schur_FlowChart}).

For the DD+PI-Full and DD+PI-Schur configurations, enforcing the dynamic equilibrium equation during training required the structural stiffness, mass, and damping matrices. Given that the problem consists of 56 spatial nodes, each with three degrees of freedom, the resulting mass, stiffness, and mapping matrices have dimensions of $168 \times 168$. To ensure a fair comparison across all configurations, we standardized the training parameters: the maximum number of epochs was set to 5,000, with a fixed learning rate of $5 \times 10^{-4}$, a batch size of 20, a training data ratio of 30\%, and an identical network architecture.

\begin{figure}[!htb]
    \centering
    \includegraphics[width=0.65\textwidth]{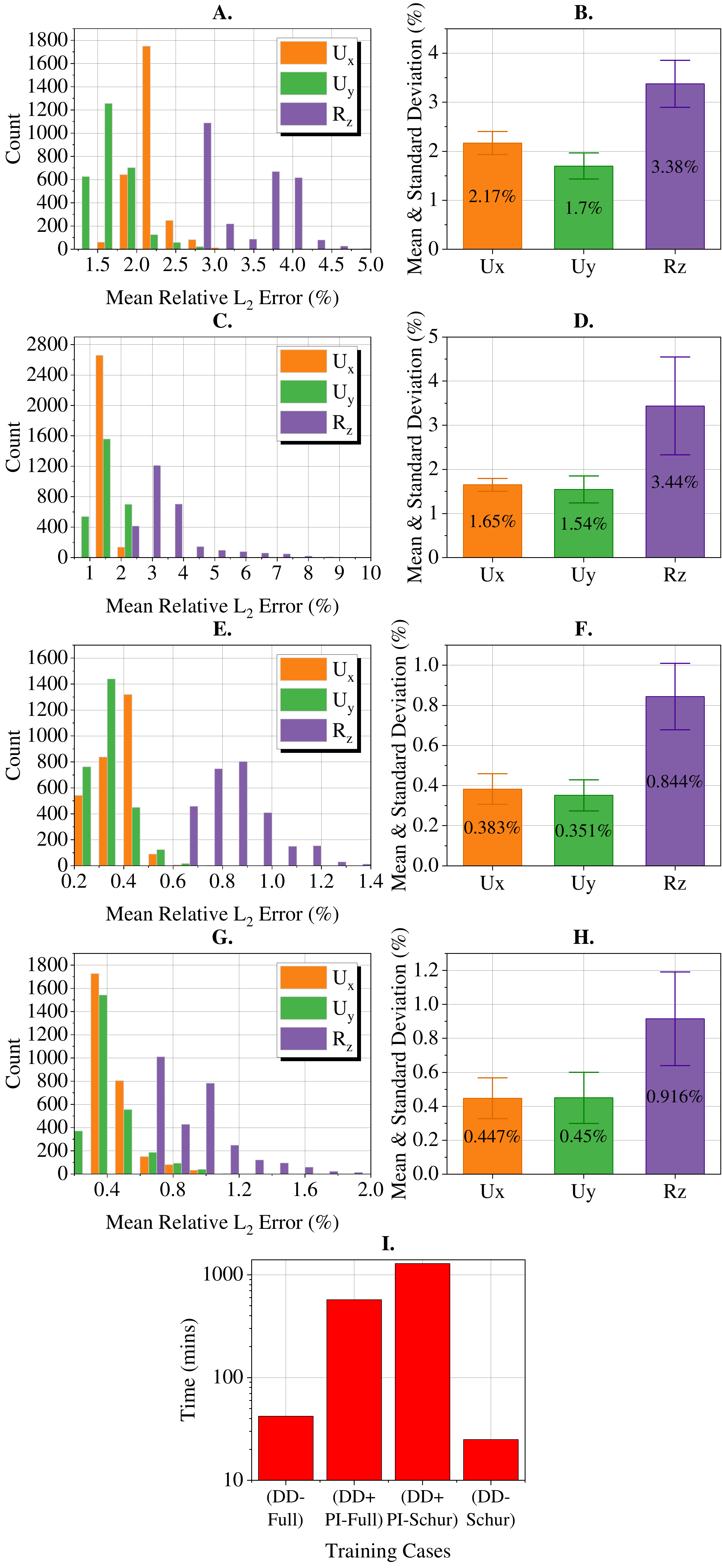}
    \caption{Prediction results for different training configurations: \textbf{A \& B.} Relative error histogram and mean-standard deviation plot for DD-Full, \textbf{C \& D.} Relative error histogram and mean-standard deviation plot for DD + PI-Full, \textbf{E \& F.} Relative error histogram and mean-standard deviation plot for DD + PI-Schur, \textbf{G \& H.} Relative error histogram and mean-standard deviation plot for DD-Schur, \textbf{I.} Training time associated with each configuration}
    \label{Results_Beam}
\end{figure}

The error histograms and training times for each configuration are presented in Figure \ref{Results_Beam}. It is evident that all configurations achieved highly accurate predictions. However, the DD+PI-Full and DD+PI-Schur configurations exhibit lower errors compared to their purely data-driven counterparts, DD-Full and DD-Schur. This confirms that integrating underlying physics enhances prediction accuracy. Despite its benefits in accuracy, the primary drawback of physics-informed training is its significant computational cost. Enforcing dynamic equilibrium requires extensive calculations to obtain acceleration and velocity at each point while ensuring consistency with the governing equations. Consequently, DD+PI-Full results in a training time nearly 14 times higher than DD-Full. Moreover, applying the Schur complement approach (DD+PI-Schur) further amplifies the computational expense, making training time almost 30 times higher than DD-Full. This is expected, as each training step necessitates solving for all system nodes based on the selected Schur nodes to satisfy Eq. \ref{Main_Eq}, making it computationally intensive and significantly more expensive.

Figure \ref{Results_Beam} also illustrates that the error for DD+PI-Schur and DD-Schur is significantly lower compared to DD-Full. This is because the network is trained on a smaller spatial domain, and the solutions for the remaining nodes are reconstructed through a post-processing technique. The DD-Schur approach is particularly advantageous when exact solutions are available at only a few points in the domain. By training the model on these selected nodes, we ensure accurate predictions at those locations, and the solution for the entire domain is then reconstructed by enforcing the underlying physics (Figure \ref{Schur_FlowChart}). This method is a powerful tool for extending partial solutions to the full domain while maintaining physical consistency.

\begin{figure}[!htb]
    \centering
    \includegraphics[width=0.65\textwidth]{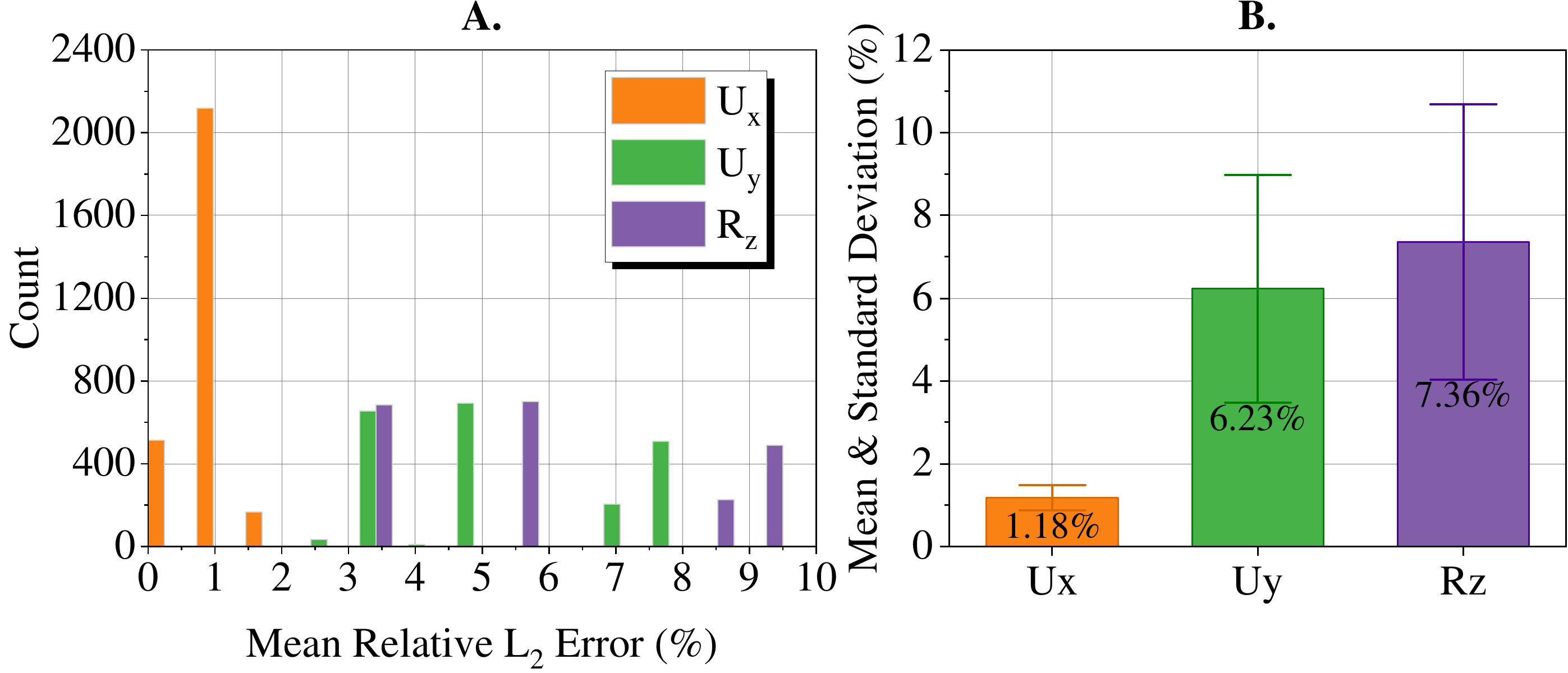}
    \caption{Post-processed results based on predictions at Schur nodes: \textbf{A \& B.} Post-processed relative error histogram and mean-standard deviation plot for the reconstructed solution over the entire domain}
    \label{Postprocess-Results}
\end{figure}

The error histogram in Figure \ref{Postprocess-Results} presents the post-processing results obtained using the Schur complement technique. It can be observed that the error is higher in post-processed results compared to purely data-driven training. This discrepancy arises because post-processing involves multiple numerical integration schemes, such as the HHT method, to obtain the solution across the entire domain. Any minor errors in the ML predictions can propagate during post-processing, leading to slightly increased overall error. Additionally, numerical integration methods inherently introduce small errors that accumulate over successive time steps, further contributing to the deviation. Nevertheless, the maximum error remains below 10\% for the rotational degree of freedom $ R_z $, which is within an acceptable range for reliable predictions across the full domain.

\begin{figure}[!htb]
    \centering
    \includegraphics[width=1\textwidth]{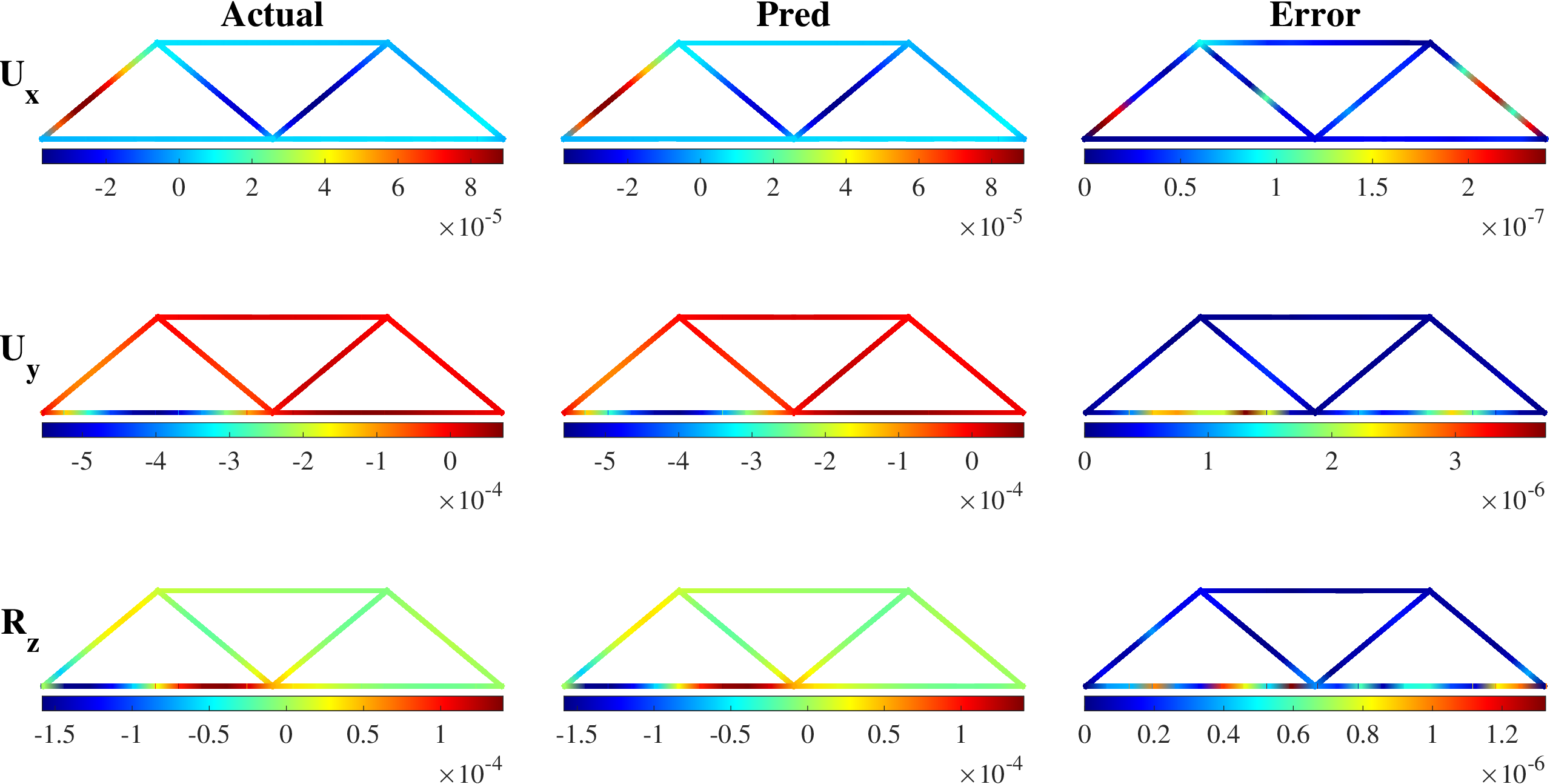}
    \caption{Actual (FEM), predicted, and error values across the entire domain for a randomly selected sample at simulation time \( T = 0.57s \): The left column shows the FEM-computed values of \( U_x \), \( U_y \), and \( R_z \), the middle column presents the predicted values of \( U_x \), \( U_y \), and \( R_z \) from the proposed network, and the right column displays the absolute error between the FEM and predicted values}
    \label{Solution_at_T0.57}
\end{figure}

\begin{figure}[!htb]
    \centering
    \includegraphics[width=1\textwidth]{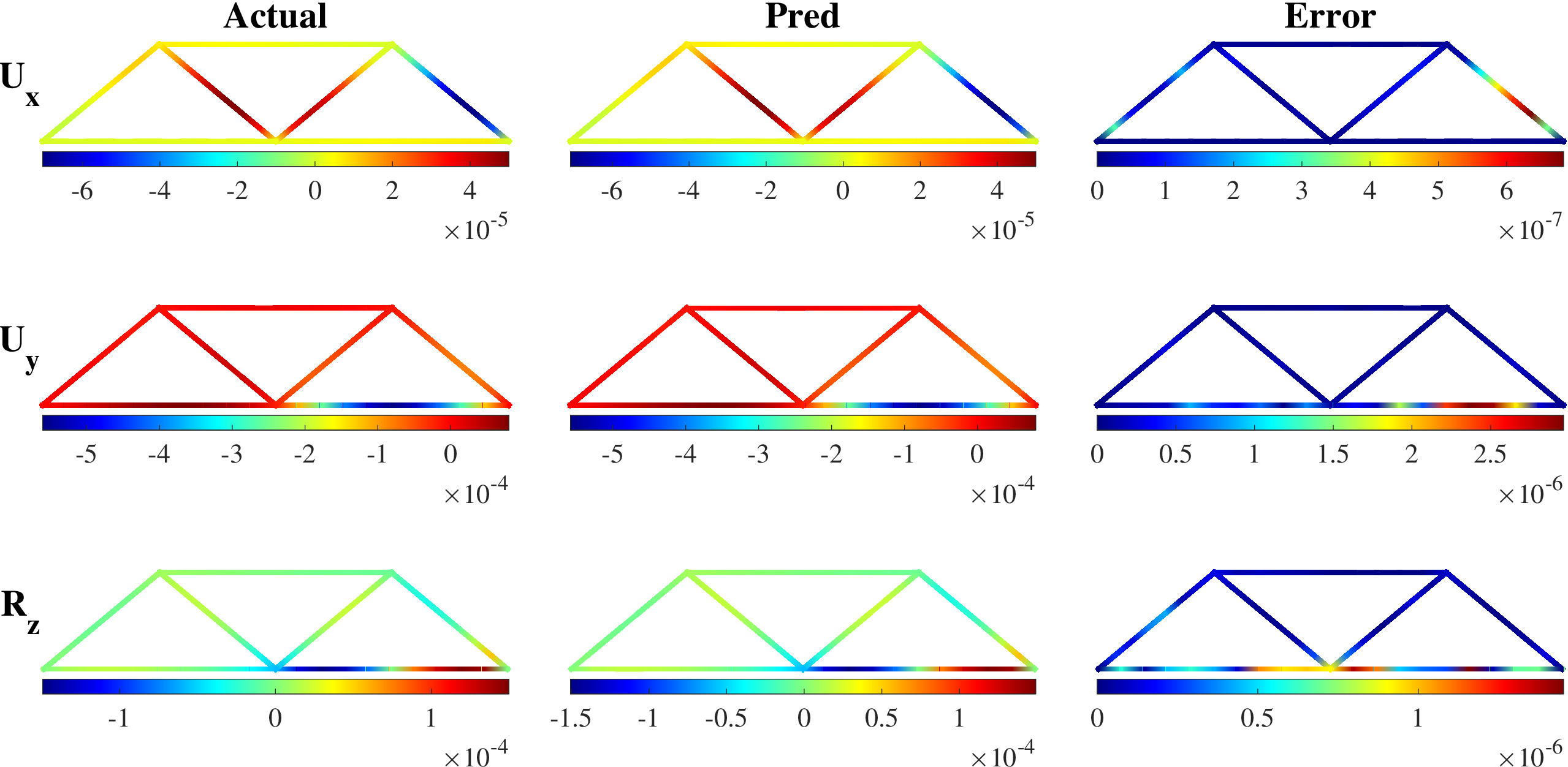}
    \caption{Actual (FEM), predicted, and error values across the entire domain for a randomly selected sample at simulation time \( T = 1.52s \): The left column shows the FEM-computed values of \( U_x \), \( U_y \), and \( R_z \), the middle column presents the predicted values of \( U_x \), \( U_y \), and \( R_z \) from the proposed network, and the right column displays the absolute error between the FEM and predicted values}
    \label{Solution_at_T1.52}
\end{figure}

Figures \ref{Solution_at_T0.57} and \ref{Solution_at_T1.52} present a comparison between the actual and predicted values of $ U_x $, $ U_y $, and $ R_z $ across the entire domain at two temporal instances: $ t = 0.57 $ seconds and $ t = 1.52 $ seconds. Additionally, the absolute error between the predicted and actual values is also visualized. The results indicate minimal prediction error, highlighting the accuracy and effectiveness of the proposed method.

\subsection{Discussion}
The results indicate that incorporating full-domain PI training significantly increases computational cost compared to purely data-driven training. The training time required for PI-based training is substantially higher due to the necessity of computing acceleration, velocity, and enforcing the dynamic equilibrium at each time step. This challenge is further enhanced when employing the Schur complement approach, where the training time becomes nearly 30 times higher than full-domain data-driven training. This makes it impractical for large-scale or real-world structural applications, especially those involving long time-series data and complex three-dimensional spatial domains.

Considering these constraints, we conclude that while PI training improves accuracy, its substantial computational cost renders it impractical for large-scale applications. Instead, we adopt data-driven training as a more feasible alternative, both for full-domain learning and Schur-domain learning. The latter enables data-driven training on a reduced set of spatial nodes while utilizing physics-informed postprocessing to reconstruct the solution across the entire domain. This strategy effectively balances computational efficiency with accuracy while ensuring that the final predictions adhere to the underlying physics. Moving forward, our work prioritizes data-driven training, supplemented by the Schur-based postprocessing framework, to obtain reliable solutions for the full domain while maintaining computational feasibility.

\section{KW-51 Bridge} \label{KW51_Sec}
\subsection{Description}
\begin{figure}[h]
    \centering
    \includegraphics[width=0.5\textwidth]{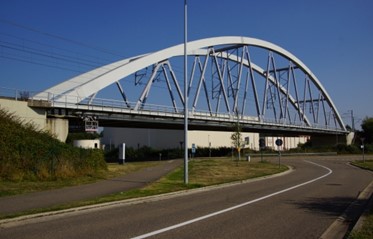}
    \caption{KW-51 railway bridge \citep{Structurae} in Leuven, Belgium}
    \label{KW51}
\end{figure}

To evaluate the proposed method on a real-life structure, we select the KW51 railway bridge in Leuven, Belgium (Figure \ref{KW51}). This steel arch railway bridge, of the bowstring type, has a total length of 115 \(m\) and a width of 12.4 \(m\). It is situated between Leuven and Brussels and features two ballasted, electrified railway tracks with curvature radii of 1125 \(m\) and 1121 \(m\), respectively.

The bridge has been continuously monitored for 15 months, from October 2018 to January 2022 \citep{KW51-Monitoring-Frequency}. A comprehensive instrumentation system was installed, comprising 12 accelerometers on the arches and deck, 12 strain gauges on the deck and diagonal members, 4 strain gauges on the rails, and two displacement sensors on the roller supports. These sensors have generated a substantial amount of real-life data, which is publicly available. Further details on the instrumentation and data collection process can be found in \citep{KW51-Monitoring-Frequency}.

\subsection{FEM Modeling and Data Generation}
The FEM of the KW51 bridge, including its structural components, element types, and tie-constraint simplifications, was developed following the methodology described in our previous work. The model is validated by comparing its natural frequencies with results from operational modal analysis (OMA) using open-source monitoring data. Full modeling details and validation results are provided in Appendix~\ref{appn:FEM}. Following validation, multiple train loading scenarios are generated by varying the number of cars, axle loads, and velocities, consistent with available monitoring data. The complete data generation and processing methodology is provided in Appendix~\ref{appn:data}.
 
\subsection{MIONet Architecture and Training}
The selected network architecture follows the same design used for the 2D beam structure, consisting of one branch and two trunks with a simple MLP-based rectangular structure. The number of hidden units and layers remains unchanged, except for the final layer of the spatial trunk. Since the structure is now 3D, it introduces additional degrees of freedom that need to be predicted—six variables instead of three. To accommodate this, the final layer of the spatial trunk is modified to have six output functions, and the number of hidden units in this layer is set to [\textit{200} $\times$ \textit{\textbf{6}}]. The batch size is reduced because the number of available samples is lower compared to the 2D beam structure. The selected batch size is 10, and the training data ratio is increased to 80\%, primarily due to the limited number of samples. As discussed in the 2D beam structure section, applying physics-informed training methods such as DD+PI-Full and DD+PI-Schur to a large, real-life structure is computationally expensive and time-consuming. Therefore, the training is restricted to two configurations: DD-Full and DD-Schur. DD-Full is a purely data-driven training approach applied to the full domain, solving for all 1882 nodes without incorporating any physics constraints. DD-Schur, on the other hand, is a hybrid approach where training is performed on a subset of nodes using available data, while physics constraints are enforced during post-processing. This ensures that the solution satisfies the underlying physics while reducing computational costs. By using DD-Schur, the training time for the smaller domain is significantly reduced while still maintaining physics consistency across the full domain. The network is trained using the ADAM optimizer with the default initializer and employs ReLU, Sin, and Tanh activation functions.

\subsubsection{Comparative Study of Activation Functions}
The choice of activation function during training plays a crucial role in capturing the response. The most commonly used activation functions are ReLU and Tanh. The ReLU activation function \citep{nair2010rectified}, defined as  
\begin{equation}
f(x) = \max(0, x)
\end{equation} 
retains positive values and sets negative values to zero. This makes it highly efficient for deep networks by mitigating the vanishing gradient problem and allowing stable backpropagation during training. However, the Tanh activation function \citep{lecun2002efficient}, given by  
\begin{equation}
f(x) = \frac{e^x - e^{-x}}{e^x + e^{-x}}
\end{equation} 
overcomes this limitation by amplifying inputs within a symmetric range of (-1,1). Unlike ReLU, Tanh preserves both positive and negative values, making it more suitable for capturing smooth, wave-like oscillatory patterns.

In parallel, the sinusoidal activation function \citep{parascandolo2016taming}, defined as  
\begin{equation}
f(x) = \sin(x)
\end{equation}
has not been widely used and is often overlooked due to difficulties encountered during training. However, some studies suggest that sinusoidal activation functions can be highly effective for specific problems. While they introduce optimization challenges due to frequent sign changes in gradients, they can also enable faster learning in certain scenarios compared to monotonic functions.

\begin{figure}[h]
    \centering
    \includegraphics[width=1.0\textwidth]{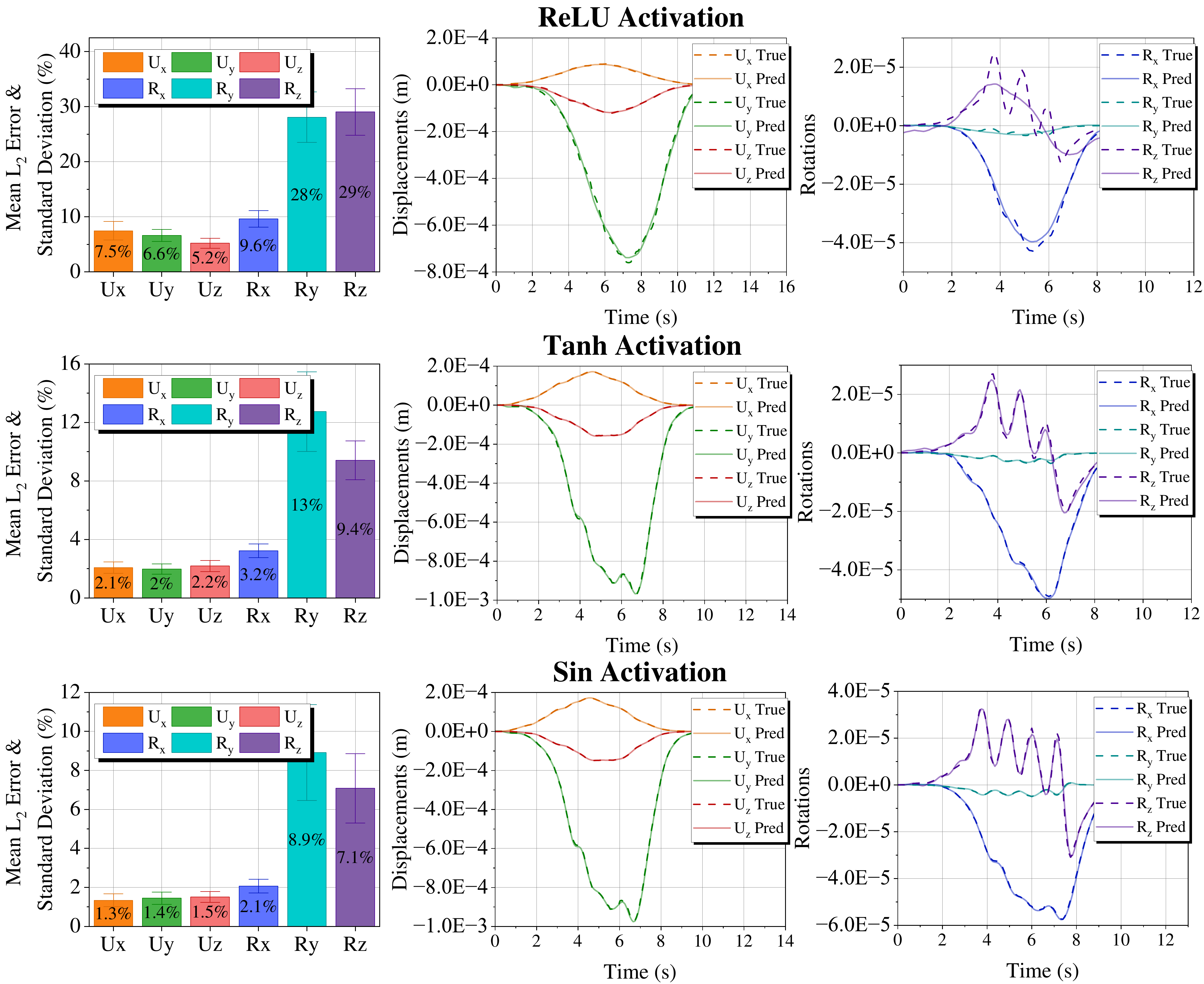}
    \caption{Comparative study of activation functions: The left column displays the mean and standard deviation of the error, the middle column compares the true and predicted displacement DOFs, and the right column compares the true and predicted rotational DOFs}
    \label{Sin_Relu_tan}
\end{figure}

To determine the most suitable activation function for this study, we test the same dataset using ReLU, Tanh, and Sin activation functions. The resulting errors and predictions for \(U_x\), \(U_y\), \(U_z\), \(R_x\), \(R_y\), and \(R_z\) at a randomly selected node across the entire time domain are shown in Figure \ref{Sin_Relu_tan}. The results indicate that ReLU fails to capture oscillatory patterns in the rotational degrees of freedom, whereas Tanh and Sin perform significantly better. Among them, the sinusoidal activation function achieves the best performance, yielding the lowest mean relative \(L_2\) error and standard deviation. Based on these findings, we select the Sin activation function for further analysis.

\subsubsection{Results}
We trained the proposed network using two configurations: DD-Full and DD-Schur. The DD-Full configuration follows a purely data-driven approach across the entire domain, using the loss function in Eq. \ref{Loss_data}. This configuration is further divided into two cases: one where the network is trained on the entire spatial mesh (1882 nodes), and the other where only the master nodes are used for training, reducing the number of training points to 998. After predictions, a post-processing step applies FEM-based constraints to obtain results for the slave nodes. The DD-Schur configuration, on the other hand, is trained on only 179 out of the 1882 spatial nodes. The loss function is given by Eq. \ref{DD-Schur_loss_Fun} without incorporating physics information during training. The responses for the remaining nodes are reconstructed through post-processing, as shown in Figure \ref{Schur_FlowChart}.

Testing the additional configurations, DD+PI-Full and DD+PI-Schur, is computationally infeasible for this problem. With 1882 nodes and six DOFs per node, the stiffness, mass, and damping matrices reach dimensions of \( 11292 \times 11292 \). Enforcing dynamic equilibrium at each training step would require multiple matrix multiplications at every time step, making it impractical. Therefore, we limit our analysis to the data-driven approaches, where the key difference is that DD-Full requires no post-processing since it is purely data-driven, whereas DD-Schur leverages a reduced training set and reconstructs the solution over the entire domain by enforcing physics information. To ensure a fair comparison, we standardized all training parameters. The maximum number of epochs was set to 5,000, with a fixed learning rate of \( 5 \times 10^{-4} \), a batch size of 10, a training data ratio of 80\%, and an identical network architecture.  

\begin{figure}[h]
    \centering
    \includegraphics[width=1.0\textwidth]{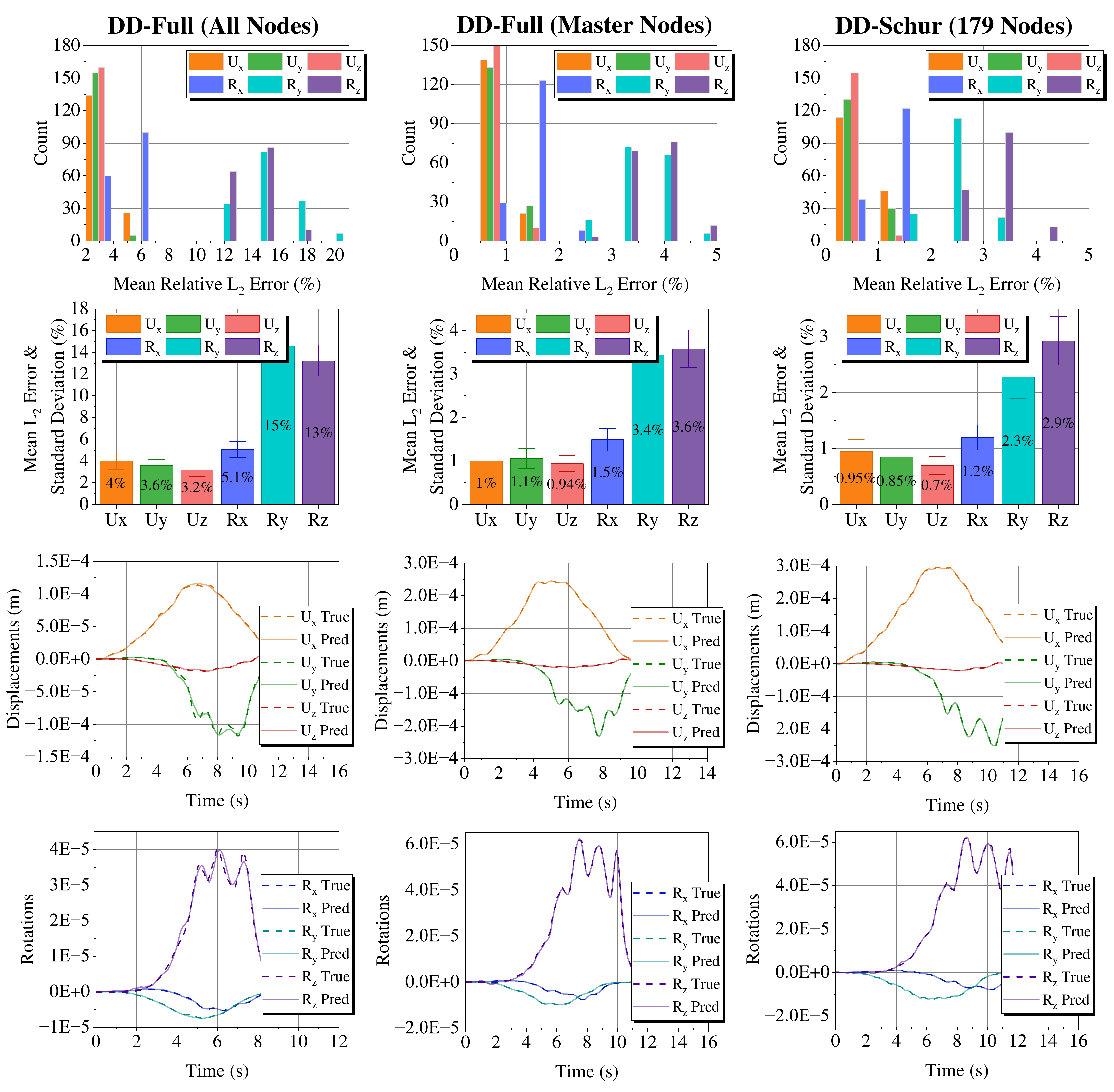}
    \caption{Prediction results for different training configurations on a random sample and node: The first row shows the relative error histogram, the second row presents the mean and standard deviation of the error, the third row compares the true and predicted displacement DOFs, and the fourth row compares the true and predicted rotational DOFs.}
    \label{histo_K51}
\end{figure}

The error histograms and variable predictions at a randomly selected node over the full time span for each configuration are presented in Figure \ref{histo_K51}. The results indicate that all configurations achieve high accuracy. As the number of spatial training nodes decreases from the full domain to master nodes and then to Schur nodes, the accuracy of the ML predictions improves. This improvement is attributed to the reduced complexity of the training data, allowing the network to learn more effectively.  

Table \ref{tab:Training_time_Kw51} summarizes the training time, inference time, and FEM simulation time for obtaining results for the same model using Abaqus. The results indicate that as the number of training nodes decreases, the training time reduces significantly. The DD-Full configuration, when trained on all nodes, requires approximately four hours, whereas limiting training to master nodes reduces it to two hours. The DD-Schur configuration further decreases training time to 0.6 hours. For inference, the DD-Full configuration produces results in 0.04 seconds per sample since no post-processing is required. In contrast, DD-Full with master nodes requires an additional post-processing step to obtain the solution for the slave nodes, which increases the inference time to two seconds. The DD-Schur configuration, requiring two post-processing steps—mapping Schur nodes to master nodes and then reconstructing the solution for the entire domain—results in an inference time of 36 seconds per sample. Although FEM provides the most accurate results, obtaining a single solution sample in Abaqus takes approximately one hour. In contrast, once the ML model is trained, the DD-Schur model can provide the solution in under one minute, demonstrating the efficiency of the ML-based approach in predicting structural responses under moving loads.  

\begin{figure}[h]
    \centering
    \includegraphics[width=0.8\textwidth]{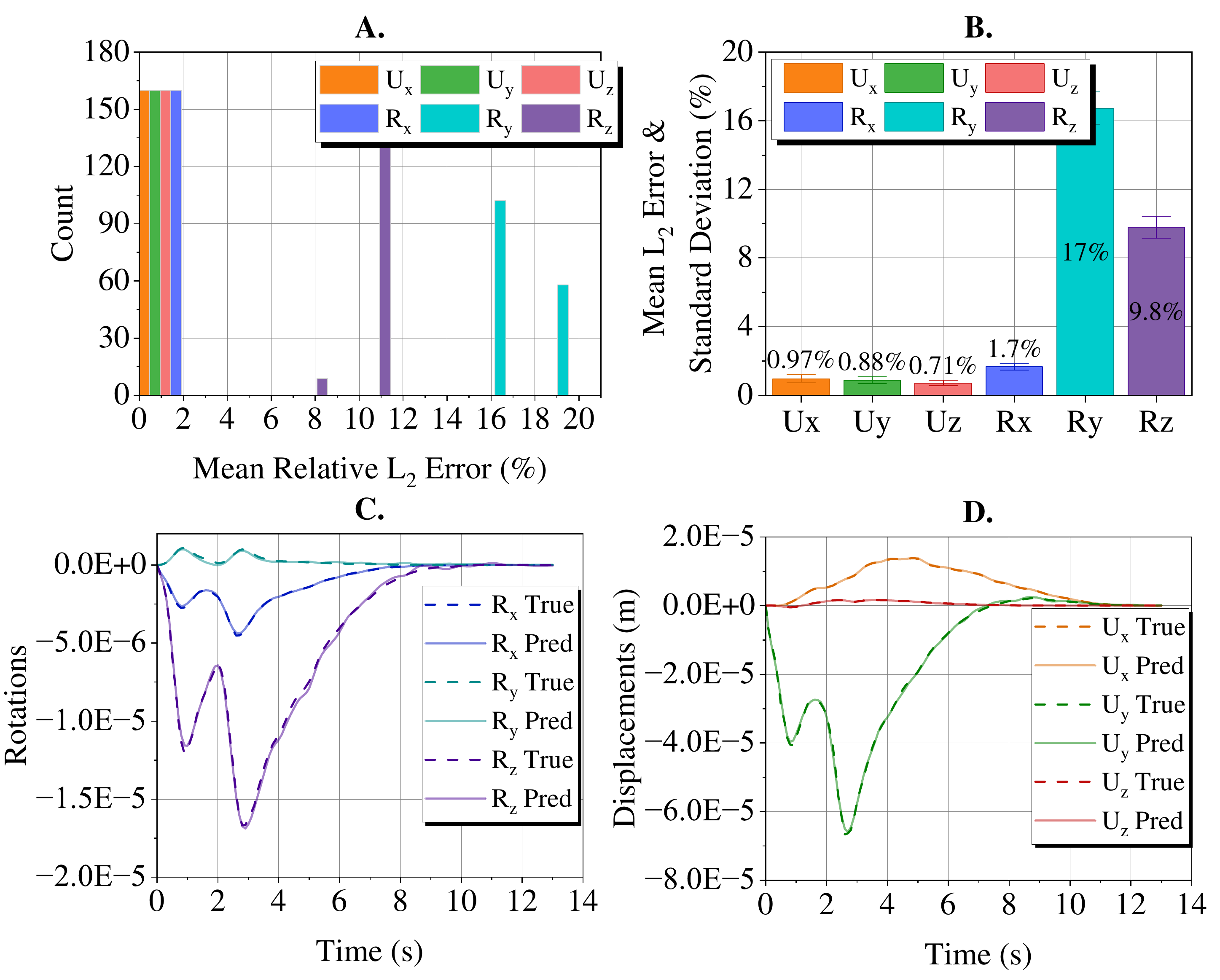}
    \caption{Post-processed results based on predictions at Schur nodes for the DD-Schur configuration: \textbf{A \& B.} Post-processed relative error histogram and mean-standard deviation plot for the reconstructed solution over the entire domain, \textbf{C.} True and predicted displacement DOFs at a random node, \textbf{D.} True and predicted rotational DOFs at a random node.}
    \label{histo_Post_K51}
\end{figure}

The DD-Schur configuration is particularly advantageous when exact solutions are available only at selected points in the domain. By training only on these nodes, accurate predictions are ensured at those locations, while the full-domain solution is reconstructed via physics-based post-processing (Figure \ref{Schur_FlowChart}). The reconstructed solution of DD-Schur configuration is shown in Figure \ref{histo_Post_K51}, where the post-processed results for major variables (DOFs) (\( U_x \), \( U_y \), \( U_z \), \( R_x \), \( R_z \)) exhibit very low errors, comparable to DD-Full (All Nodes). The only exception is \( R_y \), which shows slightly higher error due to numerical integration errors accumulating in post-processing. This discrepancy arises because the post-processing step involves multiple numerical integration schemes, such as the HHT method, which can propagate minor ML prediction errors over time. However, it is important to emphasize that \( R_y \)  represents a minor DOF, corresponding to rotation about the vertical axis, and its magnitude is extremely small—typically on the order of \(10^{-7}\) radians. For comparison, the major translational DOFs, namely \( U_x \), \( U_y \), and \( U_z \), exhibit peak values in the range of \(10^{-5}\) meters, while the dominant rotational DOFs such as \( R_x \) and \( R_z \) range between \(10^{-5}\) and \(10^{-6}\) radians. In contrast, the magnitude of \( R_y \) is one to two orders of magnitude smaller, rendering it effectively negligible in the context of structural behavior. Although the relative prediction error for \( R_y \) appears higher, the corresponding absolute error remains minimal. Consequently, this minor DOF, despite exhibiting proportionally greater error, has an inconsequential effect on the overall dynamic response of the structure.

\begin{figure}[h]
    \centering
    \includegraphics[width=0.9\textwidth]{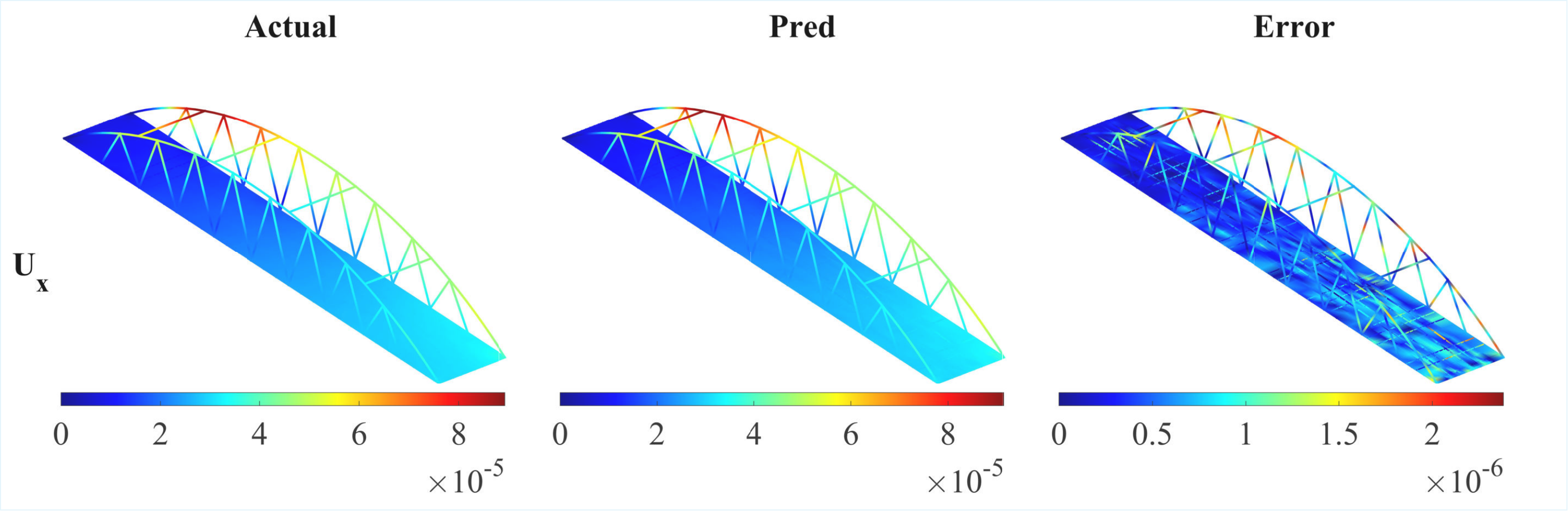}
    \includegraphics[width=0.9\textwidth]{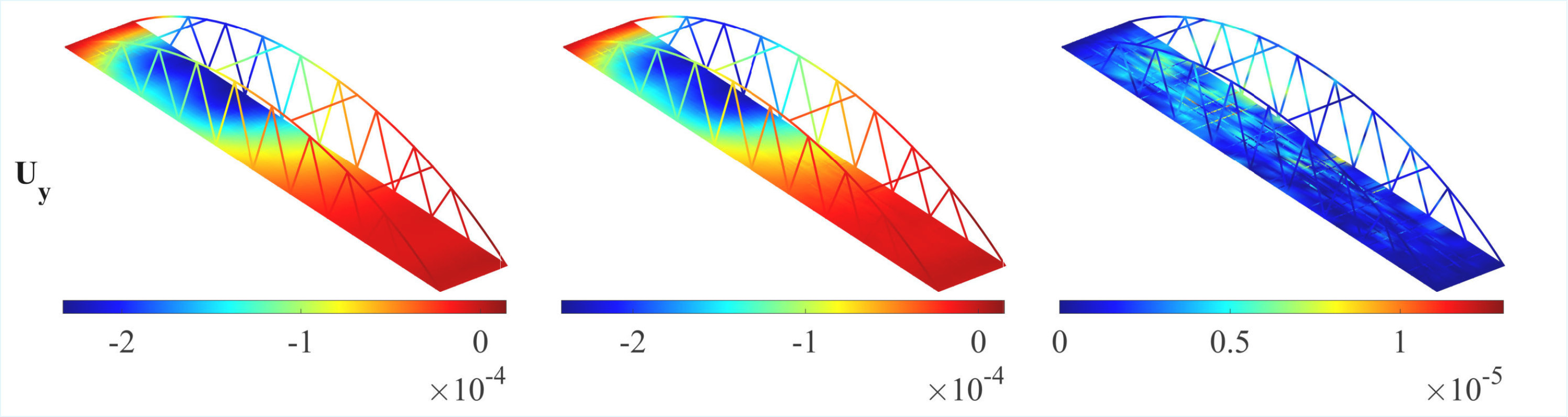}
    \includegraphics[width=0.9\textwidth]{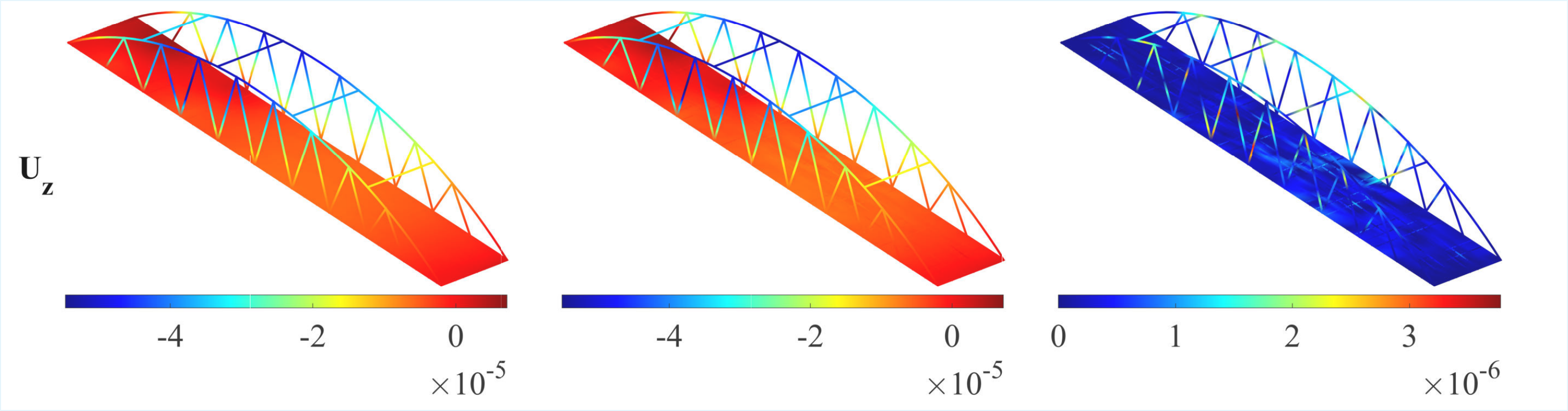}
    \includegraphics[width=0.9\textwidth]{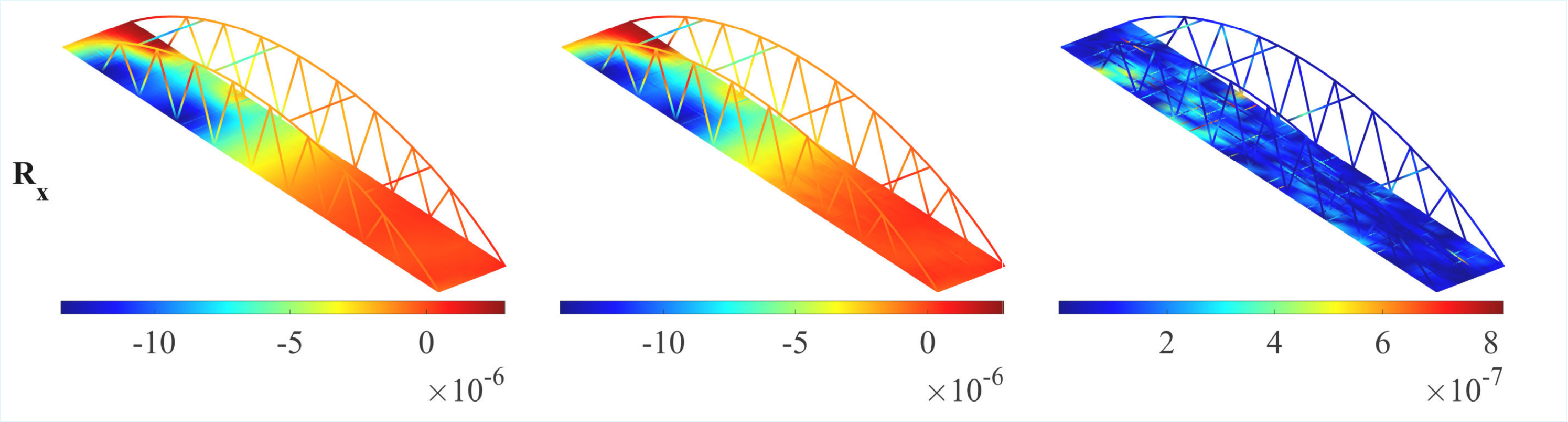}
    \includegraphics[width=0.9\textwidth]{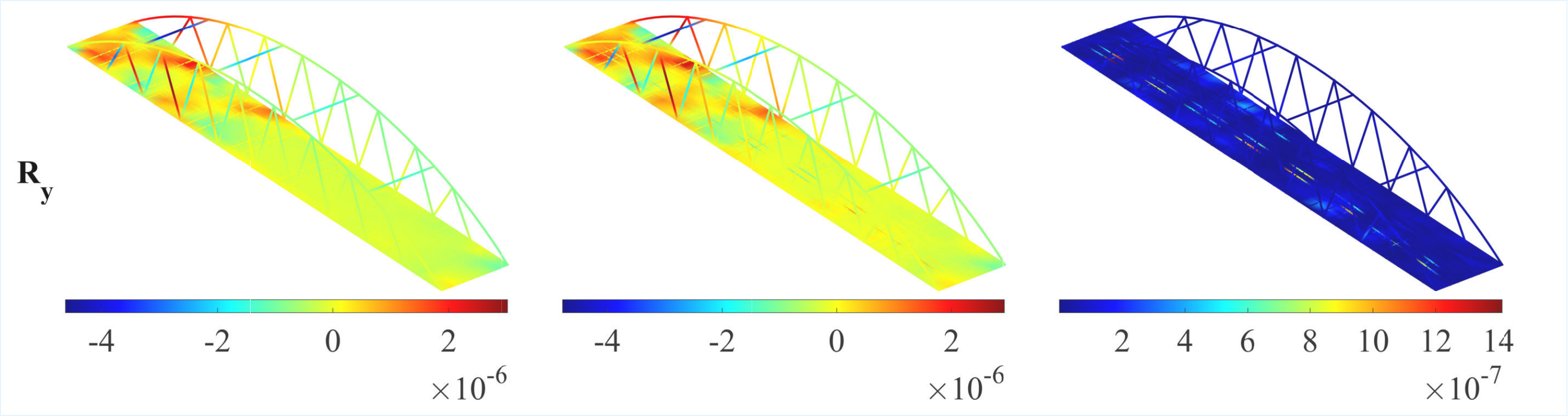}
    \includegraphics[width=0.9\textwidth]{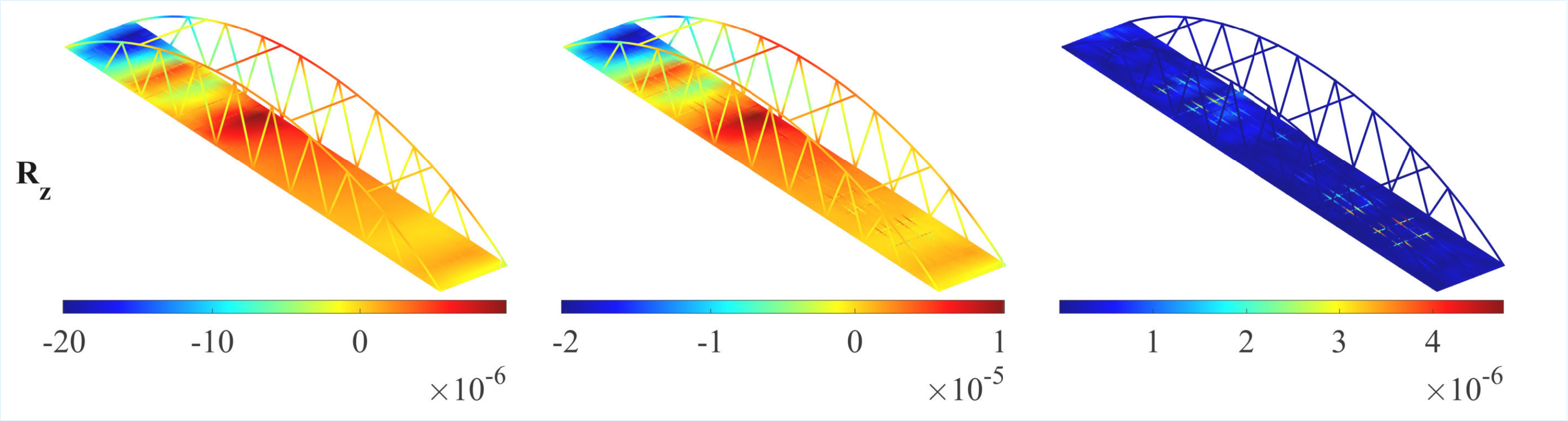}
    \caption{Actual (FEM), predicted, and error values across the entire domain for a randomly selected sample at \( T = 3 s \): Left column shows FEM values, middle column shows predicted values, and right column displays absolute errors \( U_x \), \( U_y \),\( U_z \), \( R_x \),\( R_y \), and \( R_z \)}
    \label{T=3 seconds steps 31}
\end{figure}

\begin{figure}[h]
    \centering
    \includegraphics[width=0.9\textwidth]{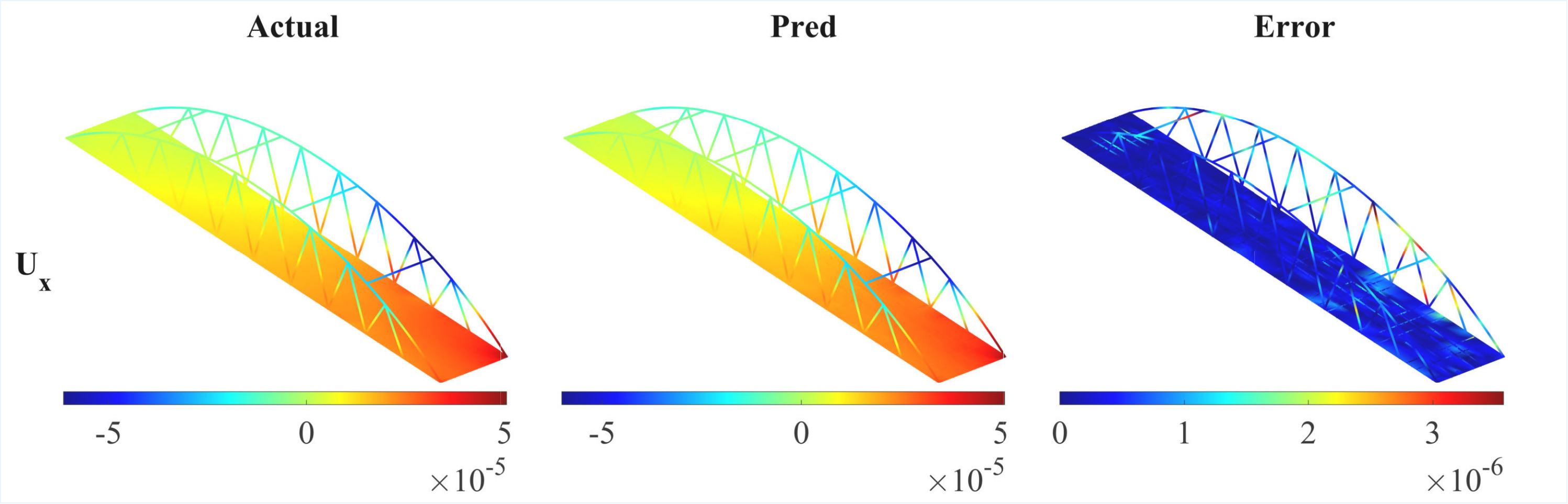}
    \includegraphics[width=0.9\textwidth]{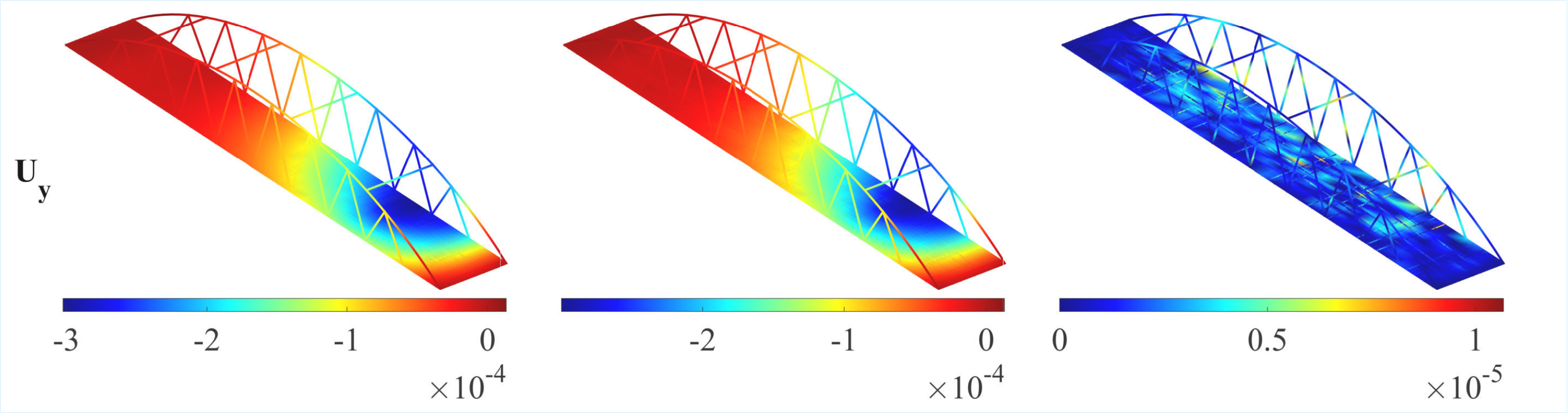}
    \includegraphics[width=0.9\textwidth]{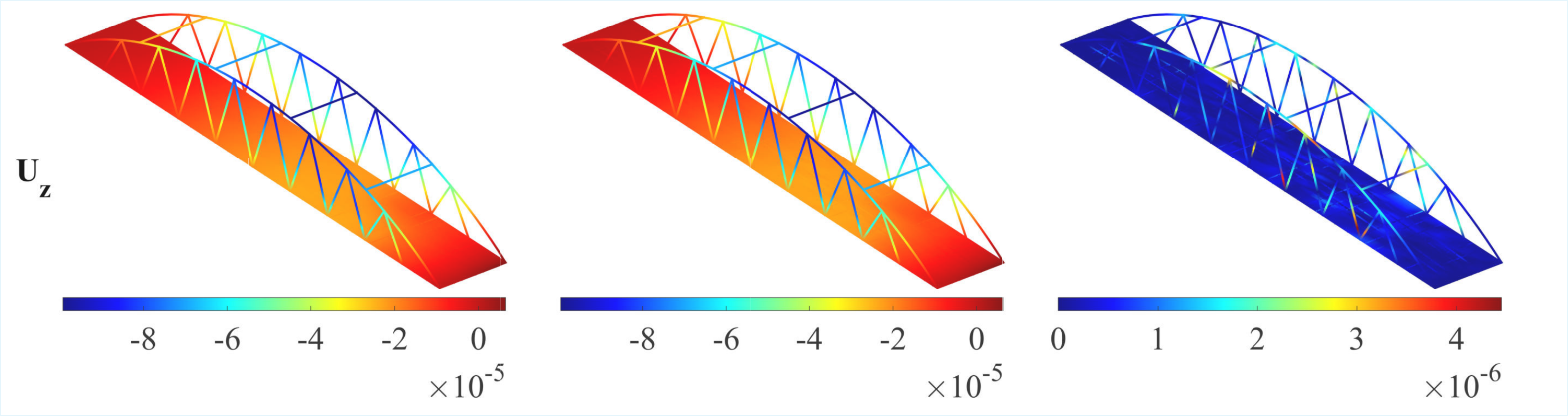}
    \includegraphics[width=0.9\textwidth]{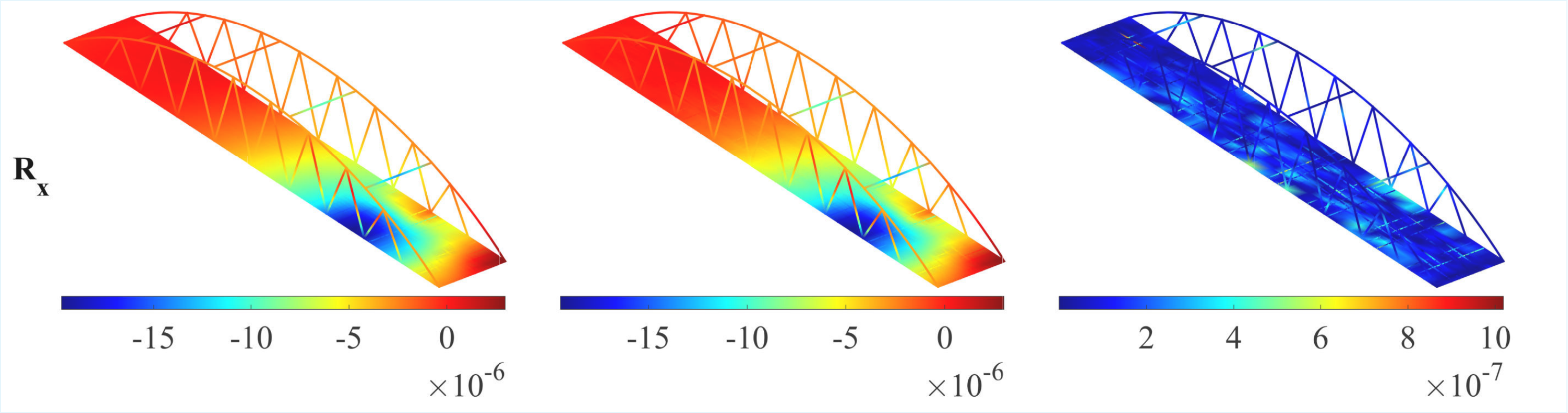}
    \includegraphics[width=0.9\textwidth]{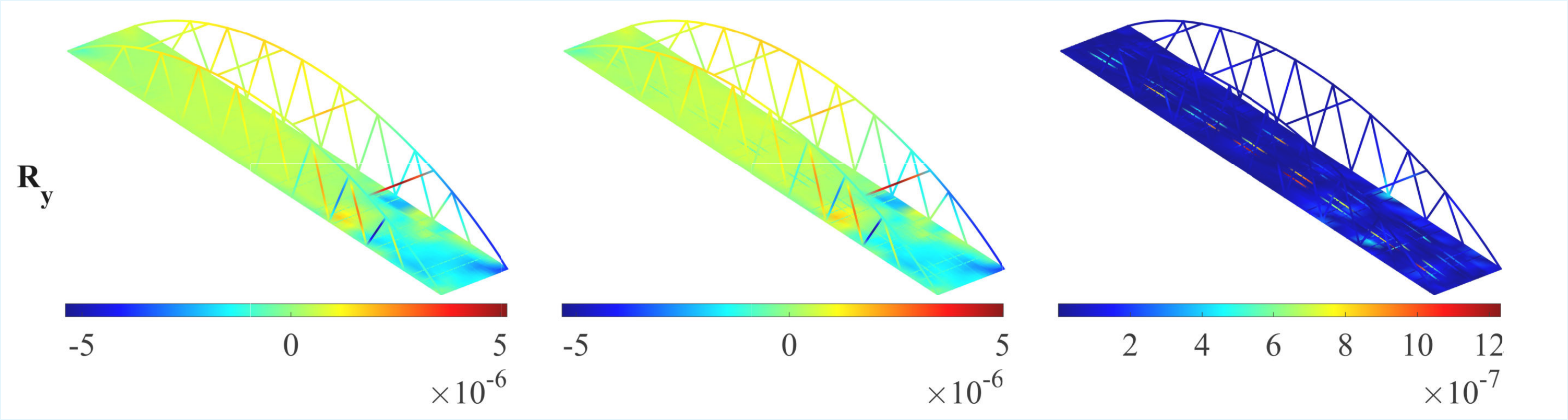}
    \includegraphics[width=0.9\textwidth]{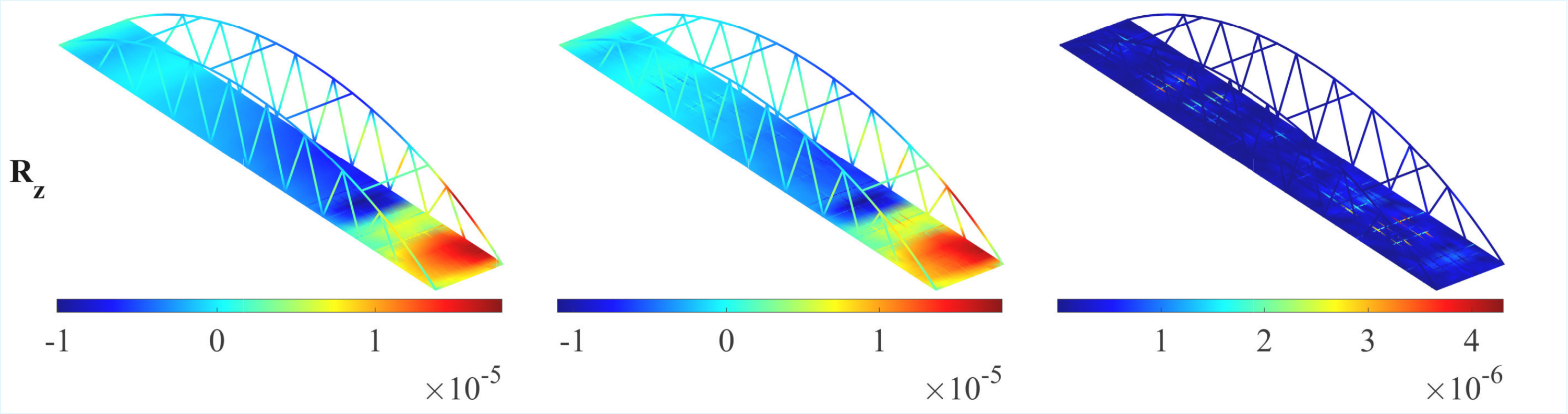}
    \caption{Actual (FEM), predicted, and error values across the entire domain for a randomly selected sample at \( T = 9 s \): Left column shows FEM values, middle column shows predicted values, and right column displays absolute errors \( U_x \), \( U_y \),\( U_z \), \( R_x \),\( R_y \), and \( R_z \)}
    \label{T=9 seconds steps 91}
\end{figure}

Figures \ref{T=3 seconds steps 31} and \ref{T=9 seconds steps 91} compare the FEM-based (actual) and predicted values for \( U_x \), \( U_y \), \( U_z \), \( R_x \), \( R_y \), and \( R_z \) across the entire domain at two time instances: \( t = 3 \) seconds and \( t = 9 \) seconds. These results demonstrate that the proposed method delivers high accuracy with minimal prediction errors, highlighting its effectiveness in approximating structural responses.

\begin{table}[]
\centering
\caption{Time comparison for different configurations}
\small
\label{tab:Training_time_Kw51}
\begin{tabular}{cccc}
\hline
\multicolumn{1}{|c|}{\textbf{Methods}}                  & \multicolumn{1}{c|}{\textbf{\begin{tabular}[c]{@{}c@{}}Training \\ (mins)\end{tabular}}} & \multicolumn{1}{c|}{\textbf{\begin{tabular}[c]{@{}c@{}}Inference/ \\ sample (mins)\end{tabular}}} & \multicolumn{1}{c|}{\textbf{\begin{tabular}[c]{@{}c@{}}Postprocessing/ \\ sample (mins)\end{tabular}}} \\ \hline
\multicolumn{1}{|c|}{\textbf{DD-Full   (Full Domain)}}  & 255                                                  & \(7 \times 10^{-4}\)                                                         & \multicolumn{1}{c|}{-}                                            \\ \hline
\multicolumn{1}{|c|}{\textbf{DD-Full   (Master Nodes)}} & 157                                                  & \(3 \times 10^{-4}\)                                                         & \multicolumn{1}{c|}{\(3 \times 10^{-2}\)}                                         \\ \hline
\multicolumn{1}{|c|}{\textbf{DD-Schur}}                 & 38                                                   & \(8 \times 10^{-5}\)                                                         & \multicolumn{1}{c|}{\(6 \times 10^{-1}\)}                                         \\ \hline
\multicolumn{1}{|c}{} & \multicolumn{3}{c|}{\textbf{Simulation Time/sample (mins)}} \\ \hline

\multicolumn{1}{|c|}{\textbf{FEM Model}}                & \multicolumn{3}{c|}{56}                                                                                                                                                                 \\ \hline
\end{tabular}
\end{table}

\subsubsection{Discussions} \label{DiscussionKW51}
The results highlight the trade-offs between different data-driven training strategies and their impact on computational efficiency and accuracy. Compared to a fully data-driven approach trained on all spatial nodes DD-Full (Full Domain), the DD-Full (Master Nodes) and DD-Schur (179 Nodes) configurations demonstrate improved accuracy while significantly reducing the training cost. This improvement is attributed to the reduced complexity of the training data, allowing the network to learn more effectively with fewer spatial points while leveraging post-processing techniques to reconstruct the full-domain solution.

A key advantage of the DD-Schur approach is its ability to achieve comparable accuracy while requiring training on only a fraction of the spatial nodes. By relying on post-processing to enforce physics-based constraints, it efficiently reconstructs the full-domain solution without the need for high-dimensional matrix operations during training. This is particularly beneficial for large-scale structural applications, where direct enforcement of equilibrium equations during training would introduce prohibitive computational costs. Additionally, the DD-Schur approach offers the advantage of training on data from specific locations (similar to placing sensors at certain points). Using the underlying physics of the system, it can then reconstruct the solution for the entire domain based on the data from those limited locations.

Despite the efficiency of the DD-Schur approach, the post-processing step introduces a minor computational overhead during inference, as it involves mapping Schur nodes to master nodes and then reconstructing the solution for the full domain. However, this additional cost remains significantly lower than that of traditional FEM simulations, where obtaining a single solution sample can take nearly an hour. The trade-off between training time and inference complexity is crucial for real-time applications, and the DD-Schur approach strikes an optimal balance by significantly reducing training time while maintaining high prediction accuracy.

\section{Conclusions, Limitations and Future Directions} \label{conclu}
This study presents an innovative framework for real-time prediction of structural dynamic responses under moving loads using MIONet. The proposed method demonstrates high fidelity in replicating FEM-level accuracy across various load magnitudes, velocities, and configurations involving multiple moving loads. This eliminates the need to rerun computationally expensive FEM simulations for each new scenario, significantly reducing analysis time and making it suitable for applications requiring real-time predictions such as digital twins.

To ensure that predictions adhere to underlying physical principles, we introduce novel PI loss functions grounded in structural dynamics. Unlike traditional PDE-based physics enforcement methods, our approach utilizes the system’s mass, damping, and stiffness matrices to formulate two dynamic loss functions: one over the full spatial domain and another based on the Schur complement for reduced domains. The full-domain loss ensures complete satisfaction of dynamic equilibrium, while the Schur-based formulation enforces equilibrium conditions within a smaller set of nodes. While these PI-based models yield improved accuracy over purely DD approaches (Section \ref{toybridge}), their computational overhead renders them impractical for real-life large-scale structures. To address this, we adopt a hybrid strategy: the model is trained purely in a data-driven manner on a reduced set of Schur nodes, and the full-domain response is then reconstructed through a physics-informed postprocessing step by satisfying the dynamic equilibrium of the system. This approach enables prediction speeds up to 100-fold faster than conventional FEM simulations, while maintaining physical consistency and predictive accuracy. Moreover, it proves particularly effective when sparse structural measurements are available, enabling domain-wide reconstruction from limited data.

Overall, the proposed method offers a powerful and computationally efficient replica to conventional FEM simulations for dynamic structural analysis. The model achieves over 95\% prediction accuracy for major DOFs, eliminates the need to rerun FE models for new loading conditions, and provides continuous output across both spatial and temporal domains. Notably, it shows promise in scenarios involving low-resolution temporal data, enabling accurate predictions at higher temporal resolutions, as demonstrated in Figure \ref{MIONet-High-Original-Low}. These capabilities position MIONet as a strong candidate for predicting real-time structural responses under complex moving load scenarios involving varying speeds and multiple loading configurations.

While the model demonstrates strong performance under elastic conditions, it assumes that no underlying changes in the structure, such as damage, aging, or material degradation, occur over time. Furthermore, its predictive accuracy is inherently tied to the fidelity of the as-built finite element model. Future developments will focus on integrating uncertainty quantification and damage-informed inputs to enhance the method’s applicability in real-world structural health monitoring.

\setcounter{section}{0}
\renewcommand{\thesection}{A.\arabic{section}}
\setcounter{table}{0}
\renewcommand{\thetable}{A.\arabic{table}}
\setcounter{figure}{0} \renewcommand{\thefigure}{A.\arabic{figure}}
\section{Parametric Study of Network Design} 
\phantomsection
\label{appn1}
\renewcommand{\thefigure}{A.\arabic{figure}}
A comprehensive parametric study is conducted to identify the optimal MLP network configuration by varying key hyperparameters, including the number of neurons per layer in both the branch and trunk networks, the number of layers, batch size, and learning rate. The proposed network architecture follows a structured approach: the branch network encodes the scaled input variables (load and velocity) and maps them to a higher-dimensional representation. The spatial trunk network processes the spatial coordinates, and its final layer is carefully reshaped to accommodate multiple output variables. In this study, three output functions—$U_x$, $U_y$, and $R_z$—are considered, necessitating that the last layer of the spatial trunk align with this dimensionality. The temporal trunk network maps the temporal domain, capturing variations over time at specified time steps. By combining the outputs of the branch and trunk networks, the model ensures a consistent prediction across the spatial domain (2D), the temporal domain (1D), and for all three output variables at each spatial and temporal discretization point. The parametric study uses the data-driven loss function defined in Eq. \ref{Loss_data}.  

\subsection{Selection of Number of Neurons}
The parametric study begins with a network architecture consisting of five layers in both the branch and trunk networks. In the spatial trunk, the final layer is designed to have neurons multiplied by the number of output functions to ensure proper mapping of all three output variables ($U_x$, $U_y$, and $R_z$). This structure remains consistent throughout the study. To maintain architectural uniformity, a rectangular network configuration is used for both the branch and trunk networks. The network performance is evaluated by varying the number of neurons per layer. Specifically, the number of neurons is tested for values in the range \{25, 50, 75, 100, 125, 150, 175, 200\}. The mean total error, computed across all output functions, is analyzed to determine the optimal configuration. As shown in Figure \ref{Parametric}\textbf{A.}, the lowest error is achieved when using 200 neurons per layer. Based on this observation, we proceed with 200 neurons in each layer for further studies.

\subsection{Number of Layers}  

After selecting 200 neurons per layer, we conducted a study to determine the optimal number of layers. The number of layers in both the branch and trunk networks was kept the same to maintain consistency. We varied the number of layers from 4 to 7 and evaluated the network performance. As shown in Figure \ref{Parametric}\textbf{B.}, increasing the number of layers improves accuracy, with the best performance observed for 6 and 7 layers. Since six layers provide a good balance between accuracy and computational efficiency, we proceed with this configuration for further analysis.  

\subsection{Batch Size}  

To analyze the impact of batch size on network performance, we tested batch sizes of 5, 10, 20, 40, and 80 while keeping the number of epochs fixed at 5000. As shown in Figure \ref{Parametric}\textbf{C.}, batch sizes ranging from 5 to 20 yield similar accuracy, whereas larger batch sizes (>20) result in a slight reduction in accuracy. Additionally, the training time (Figure \ref{Parametric}\textbf{C.}) is significantly higher for batch sizes of 5 and 10. Considering both accuracy and computational efficiency, we proceed with a batch size of 20 for further analysis.  

\subsection{Learning Rate}  

\begin{figure}[h]
    \centering
    \includegraphics[width=1\textwidth]{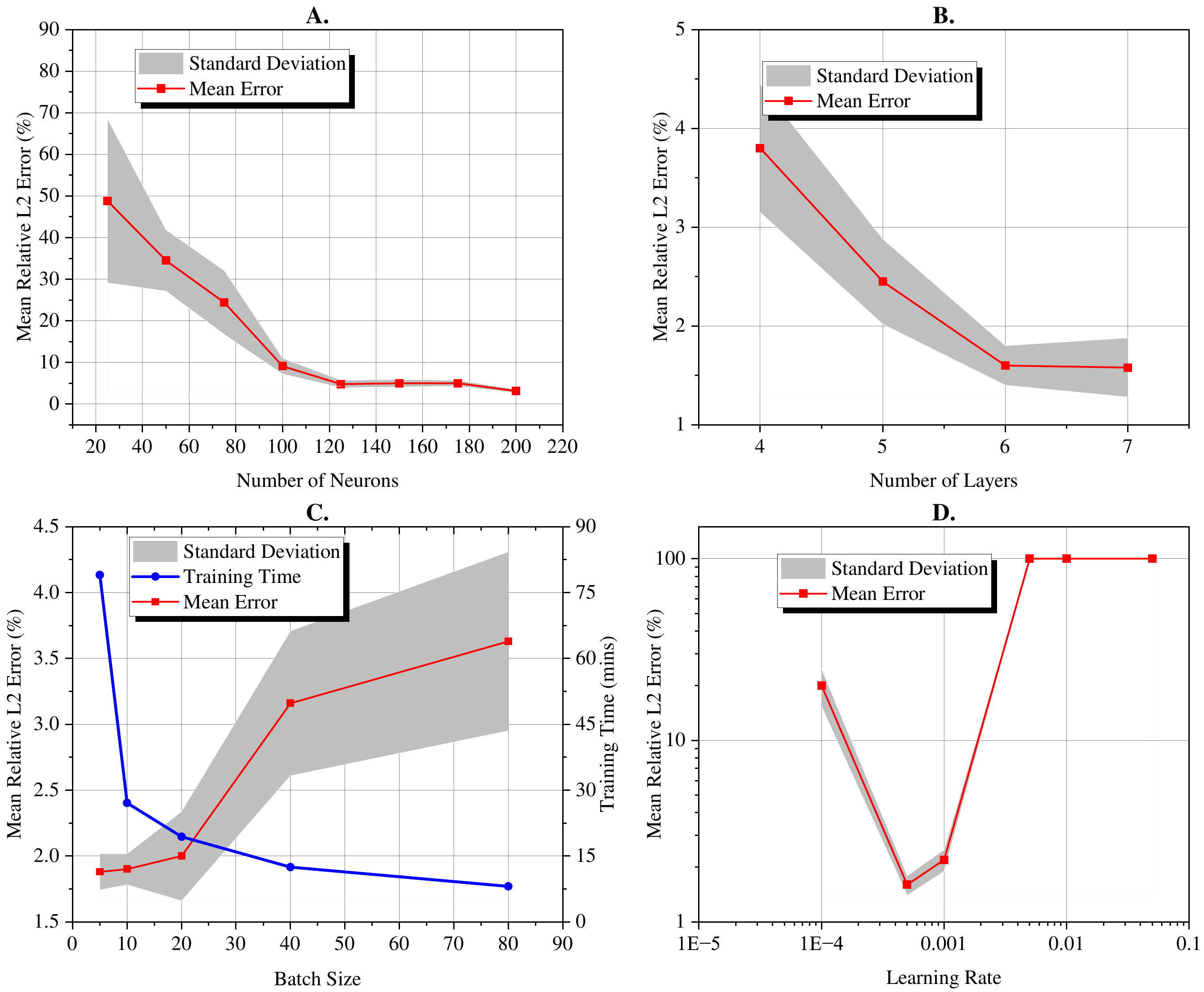}
    \caption{Parametric study: \textbf{A.} Variation of relative error with the number of neurons per layer, \textbf{B.} Relative error vs. number of layers in each network, \textbf{C.} Effect of batch size on relative error and training time, \textbf{D.} Impact of learning rate on relative error. Shaded regions show ±1 standard deviation from a single trained model.}
    \label{Parametric}
\end{figure}

We further tested the network with multiple learning rates to determine the optimal setting. As observed in Figure \ref{Parametric}\textbf{D.}, higher learning rates fail to converge and result in significantly high errors. On the other hand, lower learning rates of $5 \times 10^{-4}$ and $1 \times 10^{-4}$ exhibit similar accuracy. Therefore, we proceed with a learning rate of $5 \times 10^{-4}$ for optimal performance.

\section{KW51 FEM Modeling and Validation Details} 
\phantomsection
\label{appn:FEM}
To generate the required data, we develop a FEM of the KW51 bridge. The bridge consists of multiple components, including two arches, thirty-two diagonal members, four pipe connectors, two main girders, thirty-three transverse beams, twelve stiffeners running along the bridge length, a deck plate, a ballast layer, and two rail tracks. To simplify the modeling strategy, we model the deck plate and ballast layer using 4-node shell elements (S4R), while the remaining members are represented as 2-node Timoshenko beam elements (B31). The arches and diagonals are modeled as box sections, the girders and transverse beams as inverted T-beams, and the stiffeners as U-shaped members. The section and material properties of all components are listed in Table \ref{Parameters of KW51}.

\begin{figure}[h]
    \centering
    \includegraphics[width=1.0\textwidth]{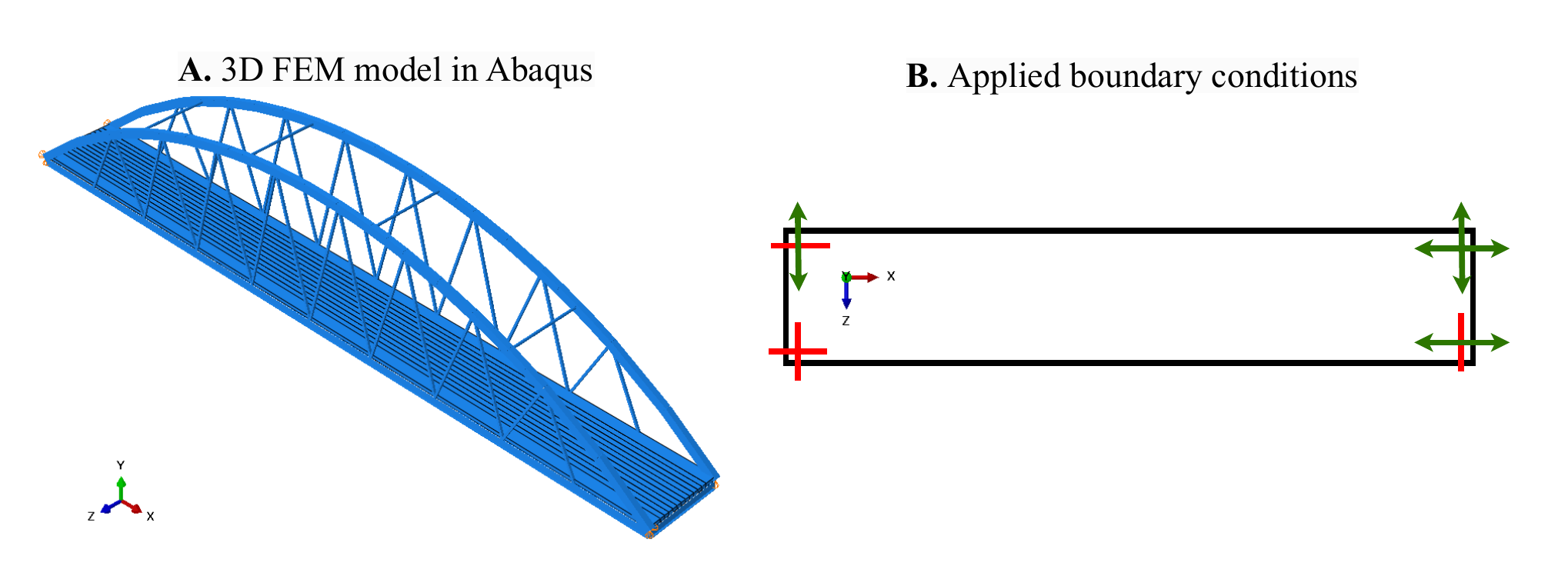}
    \caption{KW-51 modeling \textbf{A.} 3D FEM model in Abaqus, \textbf{B.} Applied boundary conditions in $xz$ plane}
    \label{KW51_modeling}
\end{figure}

To efficiently capture the dynamic elastic response of the bridge, we employ a cost-effective modeling strategy that simplifies complex structural interactions using tie constraints. The deck plate is connected to the main girders, transverse beams, and U-shaped stiffeners through tie constraints. Similarly, the ballast layer is tied to the deck plate, and the rails are tied to the ballast. Nonlinear interactions, such as friction between the deck and ballast or between the rails and deck, are excluded from this study. The actual bridge is supported by four pot bearings, which are idealized in the FEM model as pinned and roller supports (Figure \ref{KW51_modeling}). The finalized FEM model comprises 1882 nodes, each possessing six degrees of freedom, totaling 11,292 DOFs.

Once the FEM model is developed, we validate it against real-life responses by comparing its natural frequencies with those obtained through operational modal analysis (OMA). We utilize results from an OMA conducted using open-source \cite{maesmonitoring} acceleration data from sensors installed on the bridge, as reported in \citep{KW51-NaturalFrequency-MSSP}, to extract the natural frequencies of the structure. A total of 14 natural frequencies are tracked over the monitoring period based on nearly 3,000 OMA analyses. Table \ref{Natural_Freq} presents a comparison between the tracked natural frequencies and those obtained from the FEM model.

As shown in Table \ref{Natural_Freq}, the FEM model achieves over 90\% accuracy in predicting natural frequencies compared to measured values, except for the 4th and 5th modes. The discrepancy in these frequencies arises due to the unavailability of specific structural details, such as the exact dimensions of the arch box sections, connecting rods, inverted T-girders, and U-shaped stiffeners, which are not fully documented in the available literature. Although these details exist in the structure’s blueprints, contractual and security constraints prevent access. To address this, we assume values based on design guidelines and previous studies (highlighted in bold in Table \ref{Parameters of KW51}). Despite these limitations, the FEM model achieves an average accuracy of 93\% for natural frequencies, which we consider sufficient for further analysis, as validated through discussions with experts involved in bridge monitoring.

\begin{table}[]
\centering
\caption{Sectional and material properties used in FEM (Assumed values in \textbf{bold})}
\label{Parameters of KW51}
\begin{tabular}{|c|c|}
\hline
\text{\textbf{Description}} & \text{\textbf{Dimensions}} \\ \hline
\text{Length}&\text{115\(m\)}\\
\text{Width} &\text{12.4\(m\)} \\
\text{Box arch} &\text{\textbf{0.86\(m\)}$\times$1.3\(m\)$\times$\textbf{0.045\(m\)}} \\
\text{Box diagonal} &\text{0.345\(m\)$\times$0.35\(m\)$\times$0.016\(m\)} \\
\text{Pipe connector} &\text{\textbf{0.2\(m\)}$\times$\textbf{0.002\(m\)}} \\
\text{Deck thickness} &\text{0.015\(m\)} \\
\text{Ballast thickness} &\text{0.6\(m\)} \\
\text{U-shape stiffener} &\text{\textbf{0.25\(m\)}$\times$\textbf{0.25\(m\)}$\times$\textbf{0.008\(m\)}} \\
\text{T-shape girder} &\text{\textbf{0.6\(m\)}$\times$1.235\(m\)$\times$\textbf{0.08\(m\)}} \\
\hline
\text{\textbf{Description}} & \text{\textbf{Material properties}} \\ \hline
\text{Steel}&\text{$\rho$ = 7750\(kg/m^3\), \(E\) = 210\(GPa\), \(v\) = 0.3}\\
\text{Ballast}&\text{$\rho$ = 1900\(kg/m^3\) , \(E\) = 550\(MPa\), v=0.3}\\
\hline
\end{tabular}
\end{table}

\begin{table}[]
\centering
\caption{Comparison of natural frequency between the FEM and measured \citep{KW51-Monitoring-Frequency} values}
\label{Natural_Freq}
\begin{tabular}{|c|c|c|c|}
\hline
\text{\textbf{Description}} & \text{\textbf{FEM}} & \text{\textbf{Measured}} & \text{\textbf{Accuracy}}\\ \hline
\text{1$^{st}$ lateral mode of the arches}&\text{0.55}&\text{0.51}& \text{92.1\%} \\
\text{2$^{nd}$ lateral mode of the arches} &\text{1.22} & \text{1.23}&\text{99.2\%} \\
\text{1$^{st}$ lateral mode of the bridge deck} &\text{1.73} & \text{1.87}&\text{92.5\%} \\
\text{1$^{st}$ global vertical mode} &\text{2.07} & \text{2.43}&\text{85.1\%} \\
\text{3$^{rd}$ lateral mode of the arches} &\text{2.02} & \text{2.53}&\text{79.8\%} \\
\text{2$^{nd}$ global vertical mode} &\text{2.78} & \text{2.92}&\text{95.2\%} \\
\text{4$^{th}$ lateral mode of the arches} &\text{3.21} & \text{3.55}&\text{90.4\%} \\
\text{1$^{st}$ global torsion} &\text{3.53} & \text{3.90}&\text{90.5\%} \\
\text{3$^{rd}$ global vertical mode} &\text{4.04} & \text{3.97}&\text{98.2\%} \\
\text{2$^{nd}$ global torsion} &\text{4.10} & \text{4.29}&\text{95.5\%} \\
\text{2$^{nd}$ lateral mode of the bridge deck} &\text{4.52} & \text{4.81}&\text{93.9\%} \\
\text{4$^{th}$ global vertical mode } &\text{5.28} & \text{5.31}&\text{99.4\%} \\
\text{3$^{rd}$ global torsion} &\text{6.11} & \text{6.30} &\text{96.9\%}\\
\text{5$^{th}$ global vertical mode} &\text{6.34} & \text{6.83}&\text{92.8\%} \\
 \hline
\end{tabular}
\end{table}

\section{KW51 Data Generation and Processing}
\begin{figure}[h]
    \centering
    \includegraphics[width=1.0\textwidth]{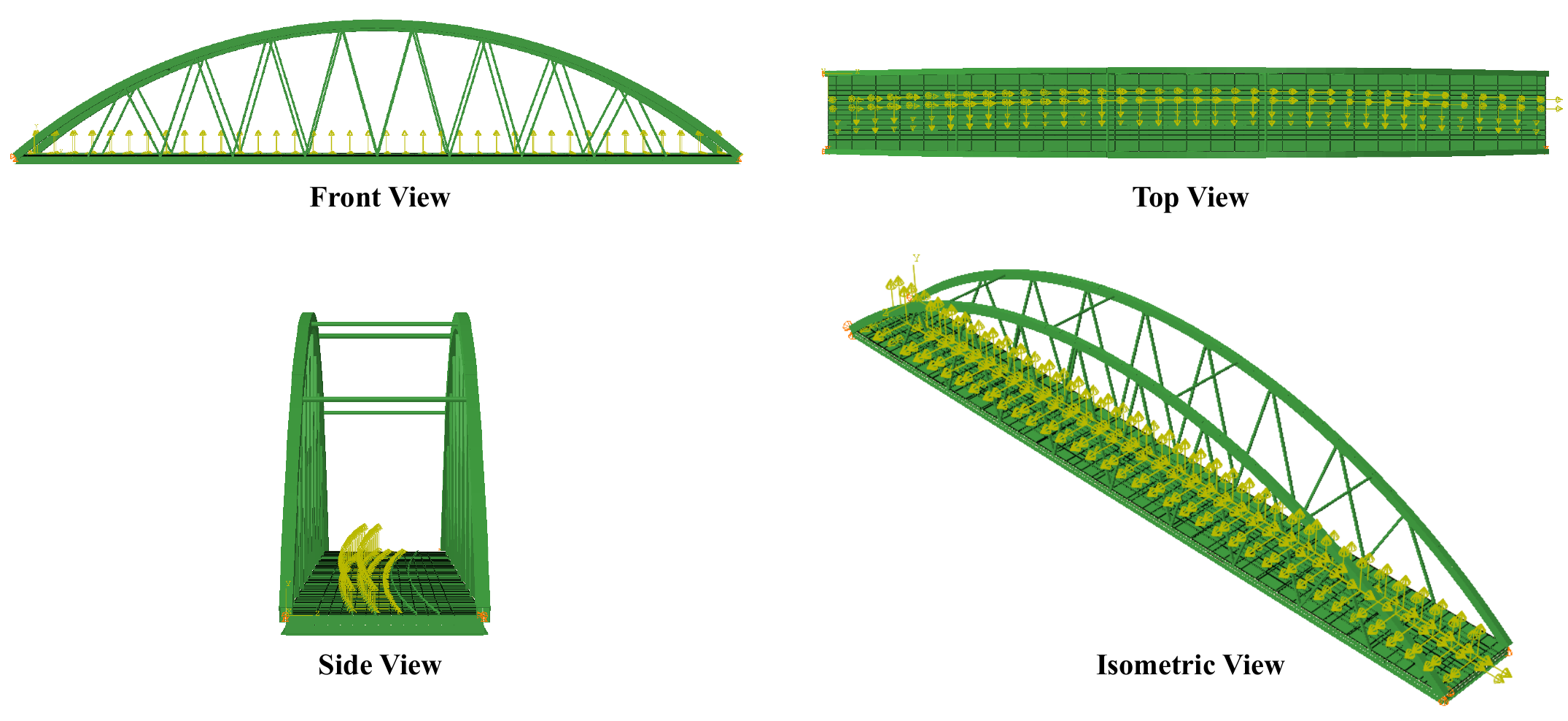}
    \caption{FEM model of the KW-51 showing the isometric, top, side, and front views to illustrate the structural configuration}
    \label{Model_view}
\end{figure}

\phantomsection
\label{appn:data}
The general views of the model are shown in Figure \ref{Model_view}, which illustrates that the track is curved. Consequently, the moving train load has a centrifugal component in addition to its vertical component. In general, the train load consists of two force components: a downward force due to gravity and an outward force due to the track curvature. The load is applied on the outer rail, which has a radius of 1125 m. 

After validating the model, we generate multiple cases by varying the load, the number of train cars, and velocity. The open-source data \citep{maesmonitoring} provides information about a train consisting of six rail cars operating at different velocities. To ensure a diverse dataset, we consider train configurations with 6, 5, 4, and 3 rail cars, leading to axle loads of 12, 10, 8, and 6, respectively. We select five velocities based on the available data: 15.75 \(m/s\), 18 \(m/s\), 21 \(m/s\), 24 \(m/s\), and 27 \(m/s\). 

\begin{figure}[h]
    \centering
    \includegraphics[width=0.7\textwidth]{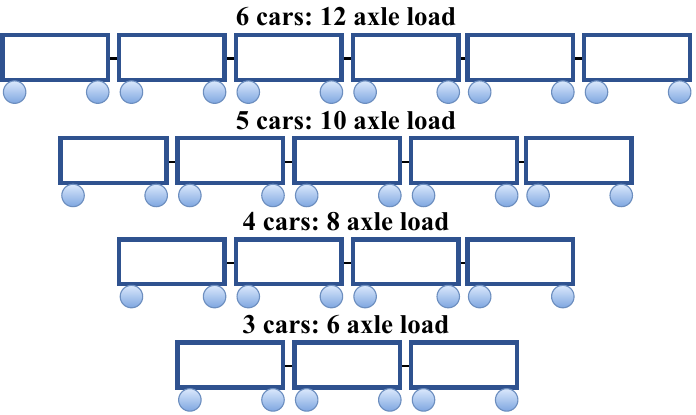}
    \includegraphics[width=0.75\textwidth]{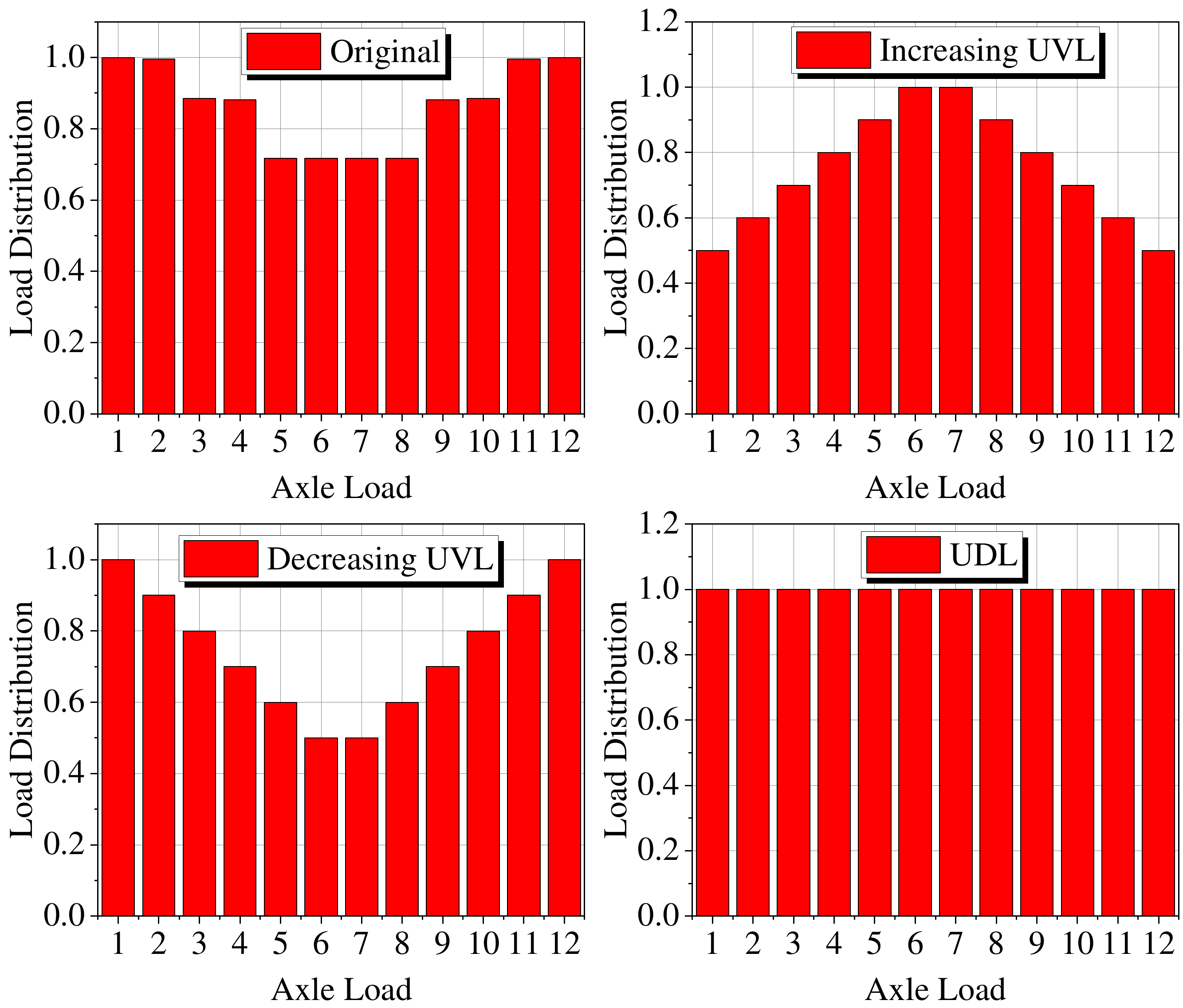}
    \caption{Train configurations used for data generation, along with corresponding load distribution on wheels}
    \label{fig:load_cases}
\end{figure}

Furthermore, the axle load distribution of the train is categorized into four distinct configurations:
\begin{itemize}
    \item Case 1: The original load distribution, as observed in real bridge testing.
    \item Case 2: Increasing uniformly varying load.
    \item Case 3: Decreasing uniformly varying load.
    \item Case 4: Uniformly distributed load across the entire train.
\end{itemize}
These four configurations, combined with the different numbers of rail cars and velocities, result in 80 unique cases. The load distribution for a 6-car train and an illustration of the number of train cars are presented in Figure \ref{fig:load_cases}. Each unique case is further simulated across multiple axle load intensities to capture a wide range of loading conditions. For every unique case, 10 samples are generated, leading to a total of 800 simulations. The primary reason for generating a limited number of samples is the high computational cost associated with each simulation. A single Abaqus simulation of the bridge under a moving load for 13 seconds, with a small time step, requires approximately one hour to complete. Additionally, post-processing the results from the Abaqus output database (ODB) to format the data for machine learning input adds further computational time. This highlights the significance of the proposed method, as it enables efficient inference using an ML model in significantly less time.

To implement the moving load in Abaqus, we define a DLOAD subroutine where multiple loads are applied at specific locations as the time step progresses. The moving load length is 0.0075 m, which corresponds to the typical contact length between a train wheel and the rail. Each simulation runs for a total duration of 13 seconds, with responses recorded at a time interval of \(\Delta t = 0.0025\) seconds, yielding 5201 time steps in the temporal dimension.

The data processing follows a similar approach to the beam structure analysis. At higher velocities and with fewer train cars, the load moves across the bridge faster, while at lower velocities and with more cars, the train stays on the bridge longer. Additionally, the number of cars influences the simulation time, which requires a stretching factor to be applied based on both velocity and the number of cars. Unlike the 2D beam structure, where a single moving load was considered, the number of moving loads here changes depending on the number of cars, which must be taken into account. To improve computational efficiency, we discard near-zero response values at the end of the simulation, similar to the previous approach. This results in varying response durations, as shown in Table \ref{tab:my-table}, leading to an uneven temporal distribution. To address this, we apply a temporal stretching technique by selecting \( v = 15.75 \ m/s \) and a 6-car train as the reference case. The responses of other cases are then scaled accordingly, introducing stretch factors \( \lambda_{v,c} \), listed in Table \ref{tab:my-table}. However, this introduces inconsistencies in the time step (\(\Delta t\)) across samples. To resolve this, we resample the response data to match the \(\Delta t\) of the reference velocity. Additionally, the total number of time steps is reduced from 5201 to 131 to minimize the temporal dimension.

Based on this, the input to the branch network becomes an \(800 \times 13\) matrix, where each sample consists of one velocity and twelve axle load values. For configurations with fewer train cars, the load input is zero for the missing axle loads. The spatial input consists of \(1882 \times 3\) matrix, representing 1882 nodes with \(x, y,\) and \(z\) coordinates. The temporal dimension consists of 131 discrete time steps, each with \(\Delta t\) of 0.01 seconds. Consequently, the final output is structured as an \(800 \times 131 \times 1882 \times 6\) tensor, where \(800\) represents the number of samples, \(131\) represents the time steps,\(1882\) represents the spatial nodes, and \(6\) represents the displacement and rotation components in the 3D domain (\(U_x, U_y, U_z, R_x, R_y, R_z\)).

% Required packages: \usepackage{multirow} \usepackage{graphicx}
\begin{table}[h]
\centering
\caption{Train simulation time and stretching factors}
\label{tab:my-table}
\small % Reduce font size
 % Scale to fit page width
\begin{tabular}{|c|c|c|c|c|}
\hline
\textbf{Velocity (m/s)} & \textbf{Cars} & \textbf{\begin{tabular}[c]{@{}c@{}}Simulation \\ Time (s)\end{tabular}} & \textbf{\begin{tabular}[c]{@{}c@{}}Time (Zeros\\ Removed) (s)\end{tabular}} & \textbf{\begin{tabular}[c]{@{}c@{}}Stretching \\ Factor ($\lambda_{v,c}$)\end{tabular}} \\ \hline
\multirow{4}{*}{15.75} & 6  & \multirow{20}{*}{13} & 13.00 & 1.000 \\  
                        & 5  &                      & 12.70 & 1.024 \\  
                        & 4  &                      & 12.10 & 1.074 \\  
                        & 3  &                      & 11.00 & 1.182 \\ \cline{1-2} \cline{4-5}  
\multirow{4}{*}{18}    & 6  &                      & 12.70 & 1.024 \\  
                        & 5  &                      & 11.50 & 1.130 \\  
                        & 4  &                      & 10.90 & 1.193 \\  
                        & 3  &                      & 10.00 & 1.300 \\ \cline{1-2} \cline{4-5}  
\multirow{4}{*}{21}    & 6  &                      & 11.30 & 1.150 \\  
                        & 5  &                      & 10.30 & 1.262 \\  
                        & 4  &                      & 9.80  & 1.327 \\  
                        & 3  &                      & 9.00  & 1.444 \\ \cline{1-2} \cline{4-5}  
\multirow{4}{*}{24}    & 6  &                      & 10.30 & 1.262 \\  
                        & 5  &                      & 9.30  & 1.398 \\  
                        & 4  &                      & 9.00  & 1.444 \\  
                        & 3  &                      & 8.20  & 1.585 \\ \cline{1-2} \cline{4-5}  
\multirow{4}{*}{27}    & 6  &                      & 9.60  & 1.354 \\  
                        & 5  &                      & 8.60  & 1.512 \\  
                        & 4  &                      & 8.30  & 1.566 \\  
                        & 3  &                      & 7.60  & 1.711 \\ \hline  
\end{tabular}
\end{table}

\section*{Credit authorship statement}
\textbf{Bilal Ahmed:} Conceptualization, Methodology, Software, Formal Analysis, Writing - Original draft, Data generation, Visualization, Supervision. \textbf{Yuqing Qiu:} Software, Formal Analysis, Writing - Review and Editing. \textbf{Diab Abueidda:} Conceptualization, Methodology, Software, Formal Analysis, Writing - Review and Editing. \textbf{Waleed El-Sekelly:} Supervision, Project administration.  \textbf{Tarek Abdoun:} Supervision, Project administration, Funding acquisition. \textbf{Mostafa Mobasher:} Conceptualization, Methodology, Writing - Review and Editing, Supervision, Project administration, Funding acquisition.

\section*{Acknowledgment}
This work was partially supported by the Sand Hazards and Opportunities for Resilience, Energy, and Sustainability (SHORES) Center, funded by Tamkeen under the NYUAD Research Institute Award CG013. This work was also partially supported by Sandooq Al Watan Applied Research and Development (SWARD), funded by Grant No.: SWARD-F22-018. The authors wish to express their gratitude to the NYUAD Center for Research Computing for their provision of resources, services, and skilled personnel. The authors would like to thank Dr. Kristof Maes (KU Leuven) for his input and advice on modeling the KW51 bridge.

\section*{Data availability}
All models and data will be made publicly available when the paper is accepted.

 \bibliographystyle{elsarticle-num} 
 \bibliography{refs}

\end{document}